# Enhancing Supply Chain Security with Automated Machine Learning

Haibo Wang
*Department of International Business and Technology Studies, Texas A&M International University, Laredo, Texas, USA*
Lutfu S. Sua
*Department of Management and Marketing, Southern University and A&M College, Baton Rouge, LA, USA*
Bahram Alidaee
*Department of Marketing, School of Business Administration, University of Mississippi, Oxford, MS, USA*

## Abstract

Global supply chains' increasing scale and complexity have led to new challenges spanning various fields, such as supply chain disruptions due to long port waiting lines, material shortages, and inflation. Coupled with the size of supply chains and availability of vast amounts of data, efforts towards tackling such challenges have led to an increasing interest in applying machine learning methods in many supply chains. Unlike other solutions, ML techniques, including Random Forest, XGBoost, LightGBM, and Neural Networks, make predictions and approximate optimal solutions faster. This paper presents an automated ML framework to enhance supply chain security by detecting fraudulent activities, predicting maintenance needs, and forecasting material backorders. Using datasets of varying sizes, results show that fraud detection achieves an 88% accuracy rate using sampling methods, machine failure prediction reaches 93.4% accuracy, and material backorder prediction achieves 89.3% accuracy. Hyperparameter tuning significantly improved the performance of these models, with specific supervised techniques like XGBoost and LightGBM reaching up to 100% precision. This research contributes to supply chain security by streamlining data preprocessing, feature selection, model optimization, and inference deployment, addressing critical challenges and boosting operational efficiency.

**Keywords:** Machine learning, supply chain management, fraud detection, material backlog, preventive maintenance

## 1. Introduction

Supply chains refer to a series of companies/organizations interacting with each other systematically to satisfy customer expectations while optimizing the use of a variety of resources. As a result of global competition in every industry, global supply chains have emerged. Consequently, managing supply chains on a global scale has become extremely complex. Meanwhile, in parallel with the advances in communication and transportation technologies, efficiency in managing the supply chains has dramatically increased recently. However, electronic data interchange among business partners has also increased the vulnerability of the supply chains against fraudulent activities. Thus, it has become essential to detect such activities to prevent fraud. Modern supply chains face high risks due to various factors. These factors include, but are not

Corresponding author: Lutfu S. Sua (e-mail: lutfu.sagbansua@subr.edu)

limited to, the size of the supplier network, the amount of financial activities involved, the volume of transactions, and information technologies with great complexity. All these factors contribute to the vulnerability of the supply chains to supply chain fraud. Moreover, this vulnerability increases with each transaction, layer, and link among the chain partners. Despite this fact, awareness of this vital issue is still not satisfactory. Association of Certified Fraud Examiners (ACFE) reports that 44.7% of the fraud incidents were discovered as a result of coincidence or a tip, while only 39.2% were found by management review, internal audit, account reconciliations, or document examination (ACFE, 2016).

Machine learning (ML) models offer great potential in solving this problem because ML models can learn from normal behavior patterns. They can also quickly adapt to deviations from normal behavior, thus identifying fraud transaction patterns. This paper implements various ML methods to detect fraud and other problems encountered in supply chains, thereby avoiding supply chain disruptions. Predictive analytics for demand forecasting, optimum routing, warehouse management, procurement, and utilization of chatbots are some supply chain applications where ML methods have been utilized successfully (Makkar et al., 2020).

Another common problem in supply chains is the prediction of material backorder. Shorter product lifecycles, the dynamic nature of global supply chains, increased competition, and many other factors make it more difficult to predict customer demand, which leads to backorders at every level of supply chains. The ability to predict the materials likely to be in short supply presents an opportunity to increase the customer service level through increased company performance (De Santis et al., 2017). Although the prediction of backorder has been considered a significant part of inventory management, backorder prediction literature is more focused on stochastic approximation (Hajek and Abedin, 2020). As a result of recent advances in machine learning coupled with the availability of a substantial amount of valuable information within the historical inventory data, ML methods can optimize the backorder decisions, resulting in higher profit values.

One essential function for supply chains with complex technology and machinery is maintenance. To avoid supply chain disruptions, preventive maintenance activities must be scheduled to ensure the continuity of operations. Preventive maintenance is simply about carrying out the required maintenance on the machines before they break down. The critical factor in preventive maintenance is the scheduling of maintenance to balance the downtime and maintenance costs. The Internet of Things (IoT) proposes a significant opportunity for companies to detect which machines or parts are likely to break through the use of connected sensors.

Through interconnected sensors on the machines, vast amounts of data are available for operators and decision makers in real time. Through such data, ML methods can analyze and learn from the patterns in the data, thus optimizing operations, predicting outcomes, and augmenting human intelligence. To help improve the security and efficiency of supply chains through the use of machine learning technology, we propose the following research questions (RQ):

RQ1: What are the accuracy levels of ML algorithms in detecting fraudulent transactions?

RQ2: Can ML algorithms improve the prediction of machine failures in supply chains?

RQ3: Can ML algorithms predict supply chain material backorders and inventory problems?

This paper proposes a data analytics framework to integrate various ML algorithms with hyperparameter tuning to find key features related to the reliability and availability of the global supply chain. Unlike other literature studies focusing on supervised learning and deep learning methods for labeled data, this proposed framework includes unsupervised and semi-supervised methods to process the unlabeled data, which is important to the online real-time system control in the supply chain. Categorical values are common in supply chain data. To the best of our knowledge, this paper is the first to compare the impact of different encoding methods on the performance of ML algorithms. This paper also addresses the issues of trustworthiness of ML algorithms by implementing Shapley values to evaluate the effect of features on prediction outcomes.

To address the challenges, this study proposes an automated ML paradigm. The flowchart in Fig. 1 illustrates this automated ML paradigm. There are four main components of automated ML. First, the data pre-processing component analyzes the characteristics of the data, provides basic information, and cleans the data for the model construction process. Second, the model construction components perform feature extraction and selection, using low code strategies to ensemble models and parameter settings. Third, an optimization process is developed in the model enhancement component to improve the trained models. Fourth, the best model/models are saved and deployed to the new data with a similar structure.

This paper is organized as follows: Related literature is reviewed in Section 2. Section 3 provides background information on the machine learning methods utilized. Section 4 compares various machine learning methods through experimental analysis and results. Finally, Section 5 provides the conclusions and potential areas for future research.

## 2. Literature Review

### *2.1 Supply Chain Security*

As vital as it is for businesses and governments to keep an eye on the disruptive supply chain disaster scenarios, it is even more important for them to devise proactive risk mitigation strategies ranging from risk analysis to building supply chain resilience. Disruption is not an unexpected phenomenon but a constant threat waiting to unfold. In this sense, COVID-19 has only revealed the vulnerability of many global supply chains to such disruptions on a scale and breadth.

Supply chain disruptions are always a potential threat due to many factors such as natural disasters, political confrontations, and even weather conditions. Companies, being aware of this fact, emphasize formulating proactive strategies to mitigate the risks by learning from past mistakes, which is only possible by analyzing past disruptions to build supply chain resiliency.

Considering the widespread impact of supply chain disruptions on national and global economies, governments need to collaborate with businesses to formulate risk management plans and develop solutions to overcome the challenges. As the challenges and risks mutate over time, companies are forced to embed resiliency into their supply chains and collaborate with the stakeholders in and out of their supply chains, including the public and private sectors, towards an up-to-date supply chain risk management model. The political, economic, and security implications of supply chain disruptions also motivate governments to develop collaboration with the private sector in a complex environment (WEF, 2012).

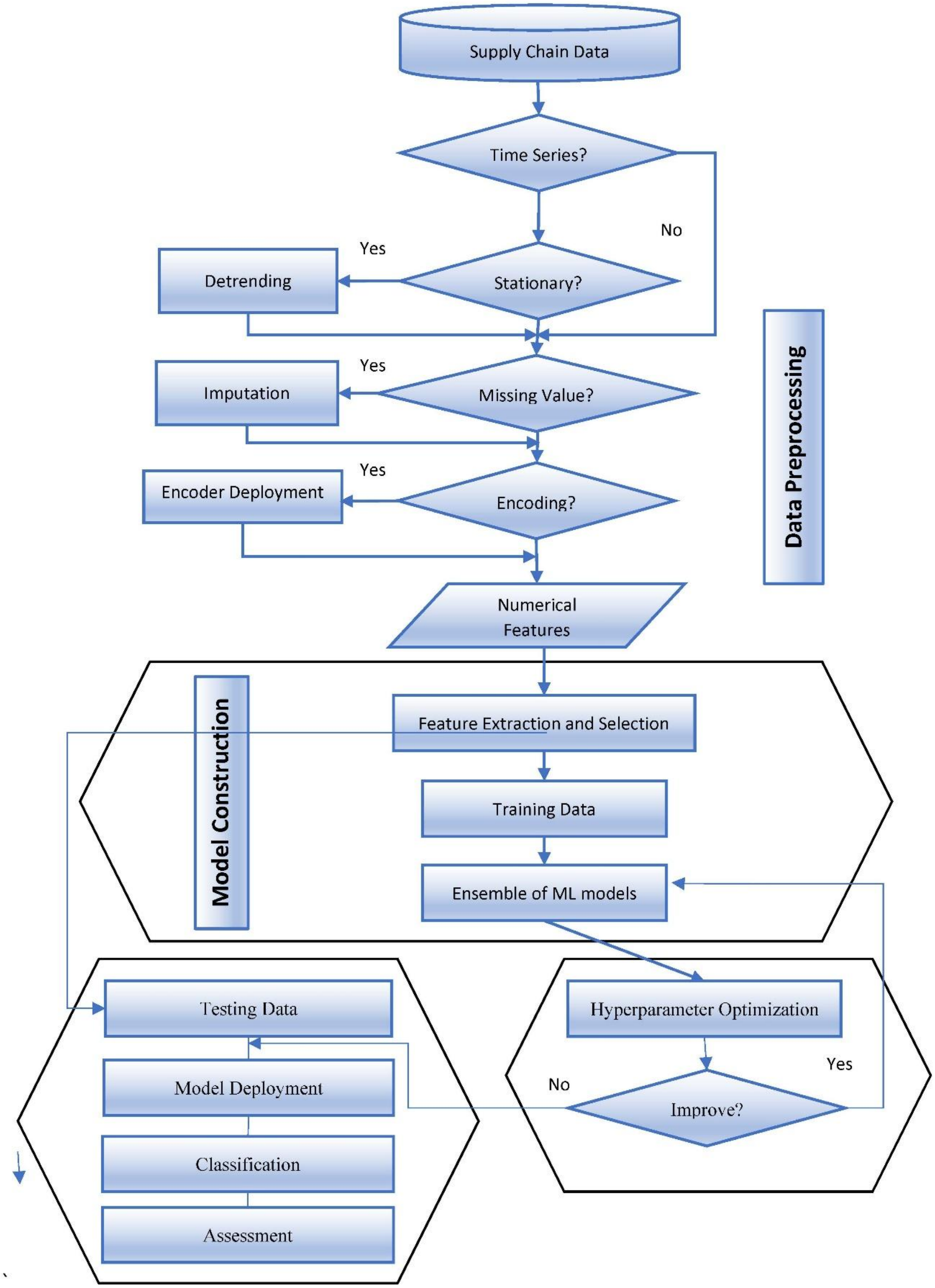

Figure 1. Model Framework

Classification of past disruptions is also important when developing risk management and response plans. Classifying supply, demand, and logistics disruptions is a possible framework for approaching supply chain challenges (Raj et al., 2022). External events causing supply chain disruptions can also be grouped under environmental, geopolitical, economic, and technological disruptions, provoking significant effects on supply chain networks (WEF, 2011). Areas where such external threats can affect supply chain vulnerabilities include, fraudulent substitution of goods causing customer dissatisfaction, incompatible security standards of the third-party service providers, and heavy dependence on the interconnected information technology systems exposing the supply chains to the single point of failure (Hassja et al., 2020) According to the most recent survey of senior decision-makers, the top three supply chain disruptions business leaders expect in 2023 are: reduced availability of raw materials in the United States, a slowdown in construction of new homes, and disruption to public transport due to lack of drivers (SAP, 2023). Increasing delivery times since the start of the pandemic are the most evident indicator of the strains in transportation of goods (Attinasi et al., 2021). Supply chain risk management (Collier and Sarkis, 2021), redesigning supply chains with blockchain-enabled circular economy (Nandi et al., 2021; Dutta et al., 2020), Business Continuity Management (BCM) for supply chain risk management (Suresh et al., 2020), and supply chain security certification (Tong et al., 2022) are some of the most recent research topics as they relate to supply chain security.

### *2.2 Machine Learning on Supply Chain Security Research*

Machine learning techniques have successfully been utilized in various aspects of supply chain management, such as demand forecasting, preventative maintenance scheduling, production scheduling, backorder prediction, cost optimization, inventory management, route optimization, and supply chain risk assessment. Table 1 summarizes recent studies involving machine learning applications in supply chain security.

Table 1. Machine learning in supply chain security

| Aspect | Details |
|---|---|
| Fraud Detection | Graph-based anomaly detection for fraud (Pourhabibi et al., 2020)<br>Use of ensemble learning for accounting fraud prediction (Bao et al., 2020)<br>Credit card default detection using ML (Srinath & Gururaja, 2022)<br>ML algorithms with quantum annealing solvers for online fraud detection (Wang et al., 2022) |
| Preventive Maintenance | Predictive models for material shortages (De Santis et al., 2017)<br>Historical sales data for identifying shortages (Malviya et al., 2021)<br>Explainable models for backorder prediction (Ntakolia et al., 2021) |
| Innovative Techniques | Graph Neural Networks for anomaly detection (Protogerou et al., 2021)<br>Distributed deep learning with Convolutional Neural Networks (CNN) for big data (Zhou et al., 2020)<br>Integration of AI with blockchain for transparency (Azzi et al., 2019) |
| Challenges Addressed | Trustworthiness of ML models<br>Comparison of ML and other forecasting methods<br>Applications in food fraud vulnerability (Yan et al., 2019)<br>Cyber resilience in supply chains (Yeboah-Ofori et al., 2022) |
| Emerging Trends | Semi-supervised and unsupervised learning for supply chains<br>Advanced optimization for hyperparameter tuning<br>Applications in Industry 4.0 systems |

Akbari and Do (2021) presented a review of machine learning literature in supply chain management and logistics. Wenzel et.al. (2019) created a reciprocal mapping to provide an overview of supply chain management-related ML methods over 10 years. Tirkolaee et.al. (2021) developed a conceptual framework to identify ML applications in supply chain management-related fields such as demand forecasting, supplier segmentation, and supply chain prediction. Detecting fraud in supply chain transactions through machine learning is one of the areas where more attention is needed. Although earlier applications have not shown a significant difference between the performances of machine learning and other forecasting methods, promising results have been reported in the literature regarding the accuracy of the results due to the advances of new ML algorithms.

Data breaches in supply chains result in significant financial losses for companies worldwide. Thus, supply chain security has led many researchers to develop various methods to tackle supply chain fraud (Yang el al., 2020; Owczarek, 2021; Fox et al., 2018). Yeboah-Ofori et al. (2022) applied ML techniques to several classification algorithms to predict threats to cyber supply chain systems and improve cyber resilience. Bao et al. (2020) used ensemble learning, a machine learning technique, in predicting accounting fraud. Srinath and Gururaja (2022) compared various machine learning methods, such as Support Vector Machine, Random Forest, XGBoost, Neural Networks, and DALEX, based on their performance metrics in an attempt to identify credit card default. Robson et al. (2020) utilized a trend analysis method to investigate the beef supply chain and its fraud vulnerability. Yan et al. (2019) used the food fraud vulnerability assessment tool to examine the perceived vulnerability to fraud for 28 firms. Schroeder and Lodemann (2021) analyzed the scientific literature on supply chain-related fields where ML methods have been implemented within supply chain risk management.

Shakibaei et al. (2023) used machine learning to design a post-disaster humanitarian supply chain to minimize human and financial losses. De Santis et al. (2017) proposed a predictive model using machine learning methods to predict material backorders. Malviya et al. (2021) compared various machine learning methods to identify parts with shortages using historical sales data. Wu and Christofides (2021) developed a predictive control scheme incorporating an ensemble of recurrent neural networks for maintenance prediction.

Several other studies have investigated machine learning and artificial intelligence applications in supply chains in recent studies. Younis et al. (2021) reviewed artificial intelligence and machine learning applications in supply chain management. Azzi and Chamoun (2019) discussed integrating blockchain technology in supply chains to increase the systems' reliability, transparency, and security. Tahereh et al. (2020) synthesized the published works on applications of Graph-Based Anomaly Detection techniques on fraud detection. Protogerou et al. (2020) developed a graph neural network approach for a distributed fraud detection scheme monitoring the complete network infrastructure. Another strategy reported to significantly reduce the processing time proposed by Zhou et al. (2020) involves implementing the distributed deep learning model of Convolutional Neural Network on a big data infrastructure.

Supervised learning algorithms, particularly the support vector machines (SVMs), have gained popularity in credit risk management applications. Liu et al. (2020) developed an ensemble SVM model for supply chain finance risk assessment. Ntakolia et al. (2021) compared several machine learning techniques in backorder prediction to solve a binary classification problem. Islam and Amin (2020) discussed the cost-effectiveness of early forecasting of backorders while utilizing a stack set machine learning methodology to predict product residues.

Matzka (2020) proposed an explanatory interface, an explainable model, and a predictive maintenance dataset. The explanatory performance of the model is evaluated and compared. Wang (2022) used supervised machine learning to detect domain names that are related to COVID-19. Abbasi et.al. (2020) proposed a methodology that uses ML in predicting solutions of large stochastic optimization models.

One of the popular techniques in machine learning is feature selection, and it impacts a model's performance to a great extent. Applying feature selection before modeling a dataset improves accuracy, reduces overfitting, and reduces the training time. Overfitting can be controlled by using hyperparameters. The model for each hyperparameter setting is evaluated to determine the optimal hyperparameters. Consequently, hyperparameters that result in the best model are chosen. The dataset itself also influences the optimal hyperparameters. Shahhosseini et al. (2022) proposed a nested algorithm that involves hyperparameter tuning while exploring the optimal weights to combine ensembles. The algorithm uses a Bayesian search to speed up the optimization process and a heuristic to obtain diverse base learners.

Other recent applications of machine learning methods in supply chain security include the use of blockchain-based supply chain management model by employing ML methods (Alshurideh et al., 2024), prediction modelling of cold chain logistics demand (He and Yin, 2021), traceability and transparency of cold chain logistics by integrating wireless sensor networks and correlation analysis (Xiao et al., 2017), transport infrastructure connectivity and conflict resolution with ML analysis (Luo et al., 2022), and predictive analytics and ML for real-time supply chain risk mitigation and agility (Aljohani, 2023).

### *2.3 Challenges of Machine Learning on Supply Chain Security*

Despite the advances in machine learning and its applicability in a long list of supply chain-related fields, ranging from demand forecasting, failure prediction, and quality inspection to warehouse optimization and distribution planning, there is a need for structured efforts aiming at improving every facet of supply chains through the efficient use of machine learning algorithms. Due to a wide range of available algorithms, it is also vital to determine the best algorithms for each supply chain application. This study aims to present an approach to applying ML to several supply chain problems and compare the performance of various ML methods on these problems.

While previous research has successfully applied supervised learning methods and ensemble models to individual supply chain problems, our proposed framework integrates supervised, unsupervised, and semi-supervised methods into a unified Automated Machine Learning pipeline. Unlike prior studies, this framework accommodates real-time unlabeled data streams, includes a comparative analysis of multiple encoding processes, and leverages SHAP values for explainability. Moreover, no prior work has comprehensively benchmarked many machine learning models across three distinct supply chain challenges (fraud detection, maintenance forecasting, and backorder prediction) using a single, automated framework with hyperparameter tuning and deployment capability.

## 3. Data and Research Method

This section provides basic information about the machine learning methods used in experimental analysis.

### *3.1 Data Collection and Analysis*

The first dataset used for the fraud detection problem includes around 180 thousand observations recorded at supply chains used by DataCo within three years. Essential preprocessing involves converting date columns into engineered features like shipping_duration and applying one-hot encoding to the numerous categorical variables. While the dataset notably lacks explicit missing values, this suggests undocumented prior cleaning, making it crucial to check for hidden placeholders. A detailed analysis of the class distribution reveals that the imbalanced fraud class is not random but is disproportionately concentrated within specific categories like Type and Market. The primary limitation of this data is its single-source bias; as it originates from one fictitious company, any model is optimized for its unique operational patterns and may not generalize to other supply chain systems without significant fine-tuning.

The second dataset utilized in the study is related to the material backorders, which are directly related to the supply chain effectiveness and service level. Machine learning methods help supply chains identify materials or parts that are likely to have shortages before they occur. The dataset comprises training and testing datasets containing variables with integer and string values. Based on the supply chain, inventory, and sales data, the algorithm's objective is to determine whether the products would go into backorder. Thus, the problem involves binary classification. The dataset requires careful preprocessing of its numerical features, often involving log transformations to correct for skewness in inventory and sales data, followed by scaling. Its handling of missing values is a key consideration, as their presence in performance columns likely signifies new products, making imputation less desirable than treating "missing" as a distinct informational category. Although the back-order class is extremely imbalanced, analysis confirms these instances correlate logically with explanatory factors like low inventory levels and high past-due stock. The dataset's core limitation is its inherent bias towards the source company's specific inventory management strategy and supplier network, meaning a model learns a particular business policy rather than a universally applicable prediction framework.

The third supply chain problem investigated is the prediction of preventive maintenance activities. A dataset of real predictive maintenance data from the industry is utilized for machine learning applications on this problem (Islam and Amin, 2020). Analysis of this dataset, which consists of sensor readings, demands significant feature engineering, such as creating rolling window statistics (e.g., mean and standard deviation of Torque) to capture trends preceding failure. While machine failure is a rare event causing class imbalance, a more nuanced analysis should confirm that distinct sensor signatures precede different failure types (e.g., Tool Wear vs. Heat Dissipation), validating the data's physical basis. The main limitation is the dataset's origin from a controlled environment with specific machines and operating loads. Crucially, its potential origin as simulated data presents a significant sampling bias, as a model's performance may not translate to the noisy and complex conditions of real-world industrial machinery. Table 2 provides more information, such as the size of each dataset and the training/testing ratio on the datasets used in this research.

Table 2. Summary of Datasets

| | **Fraud Detection**[1] | **Material Backorder**[2] | **Preventive Maintenance**[3] |
|---|---|---|---|
| Data-points | 9,567,507 | 24,117,225 | 130,000 |
| Features | 53 | 23 | 13 |
| Numerical features | 29 | 17 | 11 |

[1] https://github.com/raghav19980730/DataCo-Supply-Chain-Goods-Delivery-Prediction
[2] https://www.kaggle.com/c/untadta/data
[3] https://www.kaggle.com/code/gauravcrj/machine-predictive-maintenance/input

| | | | |
|---|---|---|---|
| Categorical features | 24 | 6 | 2 |
| Rows | 180,519 | 1,048,575 | 10,000 |
| Training Testing | | | |
| 0 (training/testing) | 105,913 /70,541 | 623,604 /415,736 | 5,784/3,877 |
| 1 (training/testing) | 2,396 / 1,666 | 5,541 / 3,694 | 21623 |
| Training (%) | 60 | 60 | 60 |
| Testing (%) | 40 | 40 | 40 |

*3.2 Coding of Data*

Machine learning methods require numerical variables. Thus, all categorical data must be encoded into numerical values before utilizing the model. There are supervised and unsupervised encoding methods. The label will be used as the baseline for the supervised encoding method to convert the categorical variables into numerical variables. For the unsupervised encoding method, no label is required for conversion. Another approach is to use an existing embedding developed from one domain to another, known as transfer learning, which is a way of employing data representation (Hancock and Khoshgoftaar, 2020). However, different encoding methods will produce several numerical features after encoding. In this study, we use a leave-one-out encoder for three datasets. Some contrast-based encoding methods use lots of memory space during the process and do not apply to datasets with a large number of features and a large number of unique categories per feature. In the meantime, some covariance-based ML methods cannot model datasets with many features after encoding due to the limited memory space. Thus, we only choose a set of encoding methods appropriate to the ML methods in Table 4 for this study. Table 4 shows the number of features with encoding methods on three datasets.

Table 3. Encoding methods

| **Encoding method** | **Library/Type** [a] | **Applicable** |
|---|---|---|
| Backward Difference | Sklearn/ unsupervised | Small-scale dataset with a few unique |
| BaseN | Sklearn/ unsupervised | Large-scale dataset with a few unique |
| Binary | Sklearn/ unsupervised | Large-scale dataset with a few unique |
| CatBoost | Sklearn/ supervised | Large-scale dataset with dynamical value |
| Count | Sklearn/ unsupervised | Large-scale dataset with high group counts for a few unique values |
| Generalized linear | Sklearn/ supervised | Small-scale dataset with a few unique |
| Hashing | Sklearn / unsupervised | Large-scale dataset |
| Helmert contrast | Sklearn / unsupervised | Small-scale dataset with a few unique |
| James-Stein | Sklearn/ supervised | Large-scale dataset with a few unique |
| Label | Sklearn/ supervised | only for encoding target labels |
| Leave one out | Sklearn/ supervised | Large-scale dataset |
| M-estimate | Sklearn/ supervised | Large-scale dataset with a few unique |
| One hot | Sklearn / unsupervised | Large-scale dataset with a few unique |
| Ordinal | Sklearn / unsupervised | Large-scale dataset |
| Polynomial contrast | Sklearn / unsupervised | Small-scale dataset with a few unique |
| Sum contras | Sklearn / unsupervised | Small-scale dataset with a few unique |
| Target | Sklearn/ supervised | Large-scale dataset |
| Weight of evidence | Sklearn/ supervised | Large-scale dataset |
| Quantile | Sklearn/ supervised | Large-scale dataset |
| Summary | Sklearn/ supervised | Large-scale dataset with a few unique |
| Wrapper | Sklearn / unsupervised | Small-scale dataset with a few unique |

[a] https://contrib.scikit-learn.org/category_encoders/index.html

Table 4. Number of features after encoding for the three datasets

| **Encoding** | **Target** | **Fraud Detection** | **Material** | **Preventive** |
|---|---|---|---|---|
| BaseN | No | 136 | 27 | 12 |
| Binary | No | 45 | 21 | 11 |
| CatBoost | Yes | 136 | 27 | 12 |
| Count | No | 26 | 21 | 11 |
| Hashing | No | 26 | 21 | 11 |
| James-Stein | Yes | 17 | 23 | 18 |
| Leave one out | Yes | 26 | 21 | 11 |
| M-estimate | Yes | 26 | 21 | 11 |
| Ordinal | No | 26 | 21 | 11 |
| One hot | No | 26 | 21 | 11 |
| Quantile | Yes | 120,086 | 89,864 | 1,941 |
| Summary | Yes | 26 | 21 | 11 |
| Target | Yes | 56 | 33 | 13 |
| Weight of | Yes | 26 | 21 | 11 |

#### *3.2.1. Variables and Measurements*

Three datasets in this study are labeled with one variable as the dependent variable or target variable, and the rest of the features are independent variables. The dependent variable has binary values; thus, we solve binary classification problems on all three datasets.

Table 5. Variables

| **Dataset** | **Dependent** | **Independent** |
|---|---|---|
| Fraud | Product status | Type, Days for shipping, Days for shipment, Benefit per order, Sales per customer, Delivery Status, Late delivery risk, Category Id, Category Name, Customer City, Customer Country, Customer Email, Customer Fname, Customer Id, Customer Lname, Customer Password, Customer Segment, Customer State, Customer Street, Customer Zipcode, Department Id, Department Name, Latitude, Longitude, Market, Order City, Order |
| | | Country, Order Customer Id, Order date, Order Id, Order Item Cardprod Id, Order Item Discount, Order Item Discount Rate, Order Item Id, Order Item Product Price, Order Item Profit Ratio, Order Item Quantity, Sales, Order Item Total, Order Profit Per Order, Order Region, Order State, Order Status, Order Zipcode, Product Card Id, Product Category Id, Product Description, Product Image, Product Name, Product Price, Product Status, shipping date, Shipping Mode |
| Backorder | Went_on_back_order | sku code, Current inventory level of component, lead_time, in_transit_qtry, forecast_x_month, sales_x_month, min_bank, potential_issue, pieces_past_due - perf_x_months_avg, local_bo_qty, General Risk Flags |
| Maintenance | Target (failed or not) | Product ID, Type, Air Temperature, Process temperature, Rotational speed, Torque, Tool wear, Failure type |

*3.2.2 Handling Imbalanced Data*

Imbalanced classes challenge machine learning algorithms as they assume a proportionate ratio of observations for each class. Imbalanced data can be observed in various fields, including fraud detection. Machine learning algorithms are developed to maximize accuracy. On the other hand, accuracy measures alone can be misleading when working with imbalanced data. Confusion matrix, precision, recall, and F1 score are the different metrics that can be used for better insight.

Table 6. Methods for Imbalanced Classification

| **Method** | **Theoretical foundation** |
|---|---|
| Undersampling-Cluster method | A sampling technique to produce a reproducible data/signal at a rate below its Nyquist rate, which is twice its upper cutoff frequency. |
| Oversampling-SMOTE method | An oversampling method for the imbalance problem aimed at balancing class distribution by increasing minority class examples by generating synthetic samples. |

A variety of algorithms are reported to perform well with imbalanced datasets. Decision trees involve learning through a hierarchy of questions, which forces both classes to be addressed. Oversampling is another method that duplicates the examples from the minority class in the training data. The undersampling method removes some observations from the majority class, which may result in losing valuable information and underfitting. Synthetic Minority Oversampling Technique, similar to oversampling, involves the generation of synthetic samples. The method utilizes the nearest neighbors algorithm to create synthetic data for model training. Sun et al. (2022) introduce a new method involving feature reduction for imbalanced data classification. The technique utilizes similarity-based feature clustering with adaptive weighted k-nearest neighbors. This study chose SMOTE to address the imbalanced datasets for ML algorithms.

*3.3. Model Construction*

*3.3.1. Feature selection and variable transformation*

This study examines three feature selection methods: LASSO, Pearson correlation, and the chi-squared method. The formulations of these methods for feature selection are given in Table 7. Based on the prediction accuracy of the initial experiment, LASSO is chosen for ML algorithms.

Table 7. Methods for Feature Selection

| **Method** | **Theoretical foundation** |
| --- | --- |
| LASSO | Lasso performs feature selection directly by setting the redundant variable coefficients to zero. $$\arg\min_{\beta_0,\beta}\left\{\frac{1}{N}\sum_{i=1}^{N}\left(y_i-\beta_0-\sum_{j=1}^{p}x_{ji}\,\beta_j\right)^2\right\}+\lambda\sum_{j=1}^{P}\lvert\beta_j\rvert$$ Thus, the generalized form of Lasso can be stated as: $$E(\beta)+\lambda R(\beta)$$ |
| Pearson Correlation | $$r=\frac{\sum(x-\bar{x})(y-\bar{y})}{\sqrt{\sum(x-\bar{x})^2\sum(y-\bar{y})^2}}$$ |
| Chi-squared | $$\chi_c^2=\sum\frac{(O_i-E_i)^2}{E_i}$$ |

After selecting a desirable set of features, machine learning is an idea that can tell you a lot about certain data and create general algorithms without the need to write code. If you support the algorithm with certain data, its logic will be based on it. Two criteria must be met for the machine learning algorithms to perform well. A sufficiently large amount of data must be available, and an actual pattern must exist within the data (Malone et al., 2020). Four types of machine learning algorithms are used within the scope of this research: Supervised learning, unsupervised learning, semi-supervised learning, and deep learning.

*3.3.2. Supervised Learning*

In this type of learning, algorithms use labeled data when making predictions based on what they have learned. Using the known information in advance, the system "learns" and interprets the data. Accordingly, the system learns from its mistakes and uses them to learn from them. Supervised learning techniques include support vector machine, logistic regression, random forests, decision trees, and Naïve Bayes classifier algorithms.

*3.3.3. Unsupervised Learning*

Unlike the supervised learning method, unlabeled and uncategorized training data is used. Unsupervised learning works by classifying and clustering data close to each other by making connections between previously untrained and unknown data. It provides inferences about the data by using the distances of the data samples from each other and the neighborhood relations. Dimension reduction and clustering are two techniques that fall under this category.

Although supervised learning methods are arguably more accurate than unsupervised learning counterparts, the data must be labeled appropriately with human intervention. These datasets allow supervised learning methods to reduce computational complexity because, in this case, they do not require large training sets to produce outcomes.

Common techniques historically associated with unsupervised learning include k-means clustering, which partitions data into clusters based on similarity; hierarchical clustering, which builds a tree-like representation of data relationships; and self-organizing maps (SOM), which map high-dimensional data to lower dimensions for visualization and clustering. These methods are widely recognized for their ability to group similar data points and uncover hidden structures in datasets.

While KNN is typically considered a supervised learning algorithm because it relies on labeled data to classify new instances, it can also be adapted for unsupervised applications. For example, in anomaly detection or density estimation, KNN is used to evaluate the proximity of a data point to its neighbors without predefined labels. This usage is often called unsupervised classification because it infers patterns directly from the data without explicit training labels. To illustrate, [Sun et al., 2022] applied KNN within an unsupervised framework to perform feature reduction in imbalanced datasets, demonstrating its versatility. However, the dual characterization of KNN may be confusing. When discussing its unsupervised use, it is critical to emphasize the context and application to avoid misinterpretation. Specifically, its unsupervised variant leverages distance metrics (e.g., Euclidean distance) to group data points or identify anomalies, independent of labeled outcomes. This nuanced understanding bridges the gap between its traditional supervised classification role and unsupervised analytical applications.

#### *3.3.4. Semi-supervised Learning*

Semi-supervised learning can be used when the input data has been partially labelled. Since it can be expensive and time-consuming to rely on domain expertise when labeling the data for supervised learning, semi-supervised and unsupervised learning can be better alternatives. Yang et al. (2022) provided a comprehensive survey on the applications of semi-supervised learning, including several supply chain applications such as pattern recognition, statistical learning, and natural language processing. Commonly used methods in semi-supervised learning include self-training, co-training, and algorithms like semi-supervised SVMs and semi-supervised autoencoders. These methods are well-suited for pattern recognition, text classification, and speech analysis applications. While CNNs are predominantly associated with supervised learning, for example, classifying images of cats and dogs as 0 or 1, they can also be adapted for semi-supervised tasks under specific circumstances. For instance, CNNs are employed in semi-supervised learning frameworks when paired with pseudo-labeling or consistency regularization techniques. In these scenarios, CNNs use a small set of labeled data to initialize training and then assign pseudo-labels to unlabeled data based on the model's predictions. These pseudo-labels are iteratively refined, enabling CNN to learn from labeled and unlabeled data. Jing et al. (2022) provided an example of a semi-supervised autoencoder CNN applied to defect detection, where the network combines labeled and unlabeled data to achieve robust predictions. However, it is important to note that CNNs are not inherently semi-supervised but require integration into specific semi-supervised frameworks to function this way. Without such adaptations, CNNs operate as supervised models.

Semi-supervised learning was incorporated to leverage our large volume of unlabeled data, improving the model's ability to detect rare security risks. This was achieved in two primary ways. The first technique, self-training (or pseudo-labeling), involves training an initial classifier on a small set of labeled data and using it to predict outcomes for the unlabeled data. The most confident predictions are then added as “pseudo-labels” to the original training set, and a final, more robust classifier is trained on this augmented dataset, allowing it to learn from a much larger pool of examples. The second approach uses anomaly detection to engineer a new feature. A model like an autoencoder is trained exclusively on unlabeled (presumably normal) data to learn its underlying structure. It then scores every transaction based on its reconstruction error—a measure of how much it deviates from the learned norm. This error score is added as a new predictive feature, providing any supervised model, such as a Random Forest, with a direct and informative signal for identifying unusual activity. Figure A2 is provided for illustration purposes.

### *3.3.5.Deep Learning*

Deep learning is a machine learning technique that utilizes artificial neural networks to enable digital systems to learn and make decisions based on unstructured, unlabeled data. These capabilities allow deep learning models to perform classification, regression, and generation tasks without explicit feature engineering (Sua, Wang, and Huang, 2025). For example, deep learning models are widely used for natural language processing, image recognition, and autonomous driving (LeCun, Bengio, & Hinton, 2015). Generally, machine learning trains artificial intelligence systems to learn by examining experiences with data, recognizing patterns, making recommendations, and adapting. Especially in deep learning, digital systems learn from examples rather than just responding to rule sets and use the information to behave similarly to humans. Table 8 lists the machine learning methods implemented in this study under each category.

Table 8. Machine Learning Methods

| **ML Method** | **Library** | **Type** | **Hyper-parameter** |
|---|---|---|---|
| Random Forest (RF) | sklearn | Supervised [a] | Yes |
| Balanced Bagging | sklearn | Supervised [a] | Yes |
| LightGBM | lightgbm | Supervised [a] | Yes |
| XGBoost | sklearn | Supervised [a] | Yes |
| Logistic Regression (LR) | sklearn | Supervised [a] | Yes |
| Linear Discriminant Analysis | sklearn | Supervised [a] | Yes |
| Naive Bayes ComplementNB | sklearn | Supervised [a] | Yes |
| Balanced Random Forest | imblearn | Supervised [a] | Yes |
| Ensemble with Decision Tree and AdaBoost | sklearn | Supervised [a] | Yes |
| Ensemble with RF and RUSBoost | imblearn | Supervised [a] | Yes |
| Neural Network MLP | sklearn | Supervised [a] | Yes |
| Ensemble with Random Tree and LR | sklearn | Semi-supervised [a] | Yes |
| Convolutional Neural Network (CNN) | keras | Deep learning [c] | No |
| Cluster-Based Local Outlier Factor-CBLOF | pyod | Unsupervised [d] | No |
| Copula Based Outlier Detector-COPOD | pyod | Unsupervised [d] | No |
| Empirical Cumulative Distribution Functions –ECOD | pyod | Unsupervised [d] | No |
| Histogram-Based Outlier-HBOS | pyod | Unsupervised [d] | No |
| IsolationForest Outlier Detector-IForest | pyod | Unsupervised [d] | No |
| Lightweight Online Detector of Anomalies-LODA | pyod | Unsupervised [d] | No |
| Minimum Covariance Determinant-MCD | pyod | Unsupervised [d] | No |
| Principal Component Analysis –PCA | pyod | Unsupervised [d] | No |
| Sampling | pyod | Unsupervised [d] | No |
| Scalable Unsupervised Outlier Detection-SUOD | pyod | Unsupervised [d] | No |

[a]https://scikit-learn.org/stable/supervised_learning.html
[b]https://scikitlearn.org/stable/auto_examples/ensemble/plot_feature_transformation.html
[c]https://keras.io/api/
[d]https://pyod.readthedocs.io/en/latest/pyod.models.html

A key strength of deep learning is its ability to generalize and learn hierarchical features from data, making it highly suitable for domains with vast, unstructured datasets. By iteratively adjusting the neurons' weights in a neural network, the model learns to map input data to desired outputs effectively. This process allows deep learning to surpass traditional tasks requiring high-level abstraction (Goodfellow, Bengio, & Courville, 2016). Deep learning has revolutionized fields like healthcare, which is used for disease diagnosis from medical images (Lundervold & Lundervold, 2019), and manufacturing, which predicts machine failures based on sensor data (Zhao et al., 2020). As deep learning evolves, its scalability and ability to handle unstructured data will likely drive advancements across diverse industries.

*3.3.6. Boosting Algorithms*

Boosting algorithms are implemented in machine learning models to strengthen accurate predictions. Boosting, which means boosting, in other words, tends to strengthen weak models. We can detect weak rules by running different algorithms from ML models to detect weak learning models. The prediction rate is also low if the correlation between two variables (features) is low. Generally, it may be wise to use a Decision Tree utilizing XGBoost, AdaBoost, and RUSBoost algorithms. The tree structure minimizes the next tree's error from the previous one, making such algorithms powerful.

*3.3.7. Bagging Algorithms*

Bagging algorithm is an ensemble learning method for creating a classifier ensemble by combining basic learning algorithms trained on different training set samples (Breiman, 1996). The main idea of the bagging algorithm is based on the principle of providing diversity by training each basic learning algorithm on different training sets. A simple random substitution sampling method is generally applied to create various training sets from the dataset. The outputs of the training sets obtained by the sampling method and the trained classification methods are combined through majority voting (Onan, 2018).

*3.4. Model enhancement*

*3.4.1. Hyperparameter Tuning of Machine Learning Methods*

Hyperparameter tuning is important in machine learning because it controls model behavior. Failure to tune hyperparameters results in producing suboptimal results. The objective of hyperparameter tuning for an algorithm $\mathrm{A}_{\lambda}$ is to find a function to minimize the expected loss $\mathcal{L}(x; \mathrm{F})$ over a sample of the dataset $\chi^{(train)}$ from an unknown natural distribution $\mathcal{g}_{\chi}$, where $\mathrm{F} = \mathrm{A}_{\lambda}(\chi^{(train)})$. The search method in practice is to find a good value for the hyperparameter. $\lambda$ to minimize error $\mathrm{E}_{\chi\sim\mathcal{g}_{\chi}}\left[\mathcal{L}\left(x; \mathrm{A}_{\lambda}\left(\chi^{(train)}\right)\right)\right]$. Thus,

$$\lambda^{(*)} = \underset{\lambda \in \Lambda}{\arg\min}\, \mathrm{E}_{\chi \sim \mathcal{g}_{\chi}} \left[ \mathcal{L}\left( x; \mathrm{A}_{\lambda}\left(\chi^{(train)}\right)\right)\right] \quad (5)$$

Lavesson and Davidsson (2006) state that hyperparameter tuning is more important than choosing the ML method. Hyperparameters of the machine learning models can be set. Hyperparameter values can be set before the start of the learning process. One way of performing hyperparameter tuning is through a randomized search on hyperparameters. This method chooses hyperparameter combinations randomly from the search space instead of evaluating every potential combination. Thus, the search is not guaranteed to find the best results. On the other hand, the method requires significantly less computational time, making it attractive. Exhaustive grid search is an alternative method that evaluates every combination of hyperparameter sets. Although it requires more computational time and effort, this method can find the best values in the grid space. Bayesian optimization is another method that results in more accurate solutions requiring less computational time. The method can also track previous evaluation models used to form a probabilistic model mapping hyperparameters to a probability of an objective function value. Weets et.al. (2020) have proposed a methodology for empirically determining the significance of tuning hyperparameters. In this study, we implement GridSearch and Bayesian Optimization for hyperparameter tuning. We choose the set of parameters sensitive to the prediction accuracy for each model.

### *3.4.2. Deployment of inference and practical use*

Machine learning methods can be deployed in various industries to solve multiple problems. Manufacturing systems can deploy ML algorithms to improve their productivity through process improvement. The ability to analyze numerical and categorical data from the production machines enables the ML methods to predict the likelihood of failures before they occur, saving potentially high maintenance and repair costs while avoiding disruptions in the supply chains. The ability of these methods to process large amounts of data makes it possible to monitor the processes for variations and react if necessary, thus avoiding potential quality defects.

Such capabilities of ML algorithms contribute to the versatility of the overall system in which they are deployed. In industry 4.0 applications that rely on automating processes in dynamic environments where variations need to be accounted for, ML methods become an integral part of the processes.

## 4. Findings

This section provides a comparative analysis of various ML methods using AUROC and Precision-Recall curves. AUROC stands for "Area Under the Receiver Operating Characteristics" and high area values under the curve indicate better method performance. If the curve approximates the 50% diagonal line, it suggests that the model randomly predicts the output variable. The AUROC curve helps visualize how well an ML classifier is performing.

The ratio of precision is calculated by dividing the number of true positives by the total of false and true positives. Precision measures the model's accuracy in predicting the positive class. It is also known as the positive predictive value. On the other hand, dividing the number of true positives by the total number of true positives and false negatives results in a ratio known as Recall. It is also known as sensitivity. Precision and recall should be considered together when an imbalance exists in the observations between two classes. A system with high precision and recall values will yield many results that are labeled correctly. On the other hand, a system with low precision and high recall values returns many results; however, most of the predicted labels are not correct compared to the training labels. On the other hand, a system with low recall and high precision returns very few results. Most of the predicted labels are correct compared to the training labels.

### *4.1. Fraud Detection*

Since supply chains must detect fraudulent activities in real time, computational time is just as important as the accuracy of the algorithms. The study achieved significant improvements in detecting fraudulent supply chain transactions by leveraging advanced ML algorithms. Techniques such as XGBoost, Ensemble Methods, and Random Forest provided near-perfect accuracy (up to 100% with tuning) with minimal false positives and negatives. Using SHAP values added interpretability, identifying key fraud indicators such as "Type" and "Days for Shipping". The AutoML framework's ability to process and analyze vast datasets in real-time ensures timely fraud detection, preventing financial losses and enhancing trust in global supply networks. A "critical ratio" compares the machine learning methods used in this research to indicate time and accuracy.

$$\textit{Critical Ratio} = (\textit{Overall Accuracy}) / \textit{Time (combined time in seconds)} \quad (6)$$

Fig.A1 illustrates the values of the critical ratio for the machine learning methods with and without hyperparameter tuning, where the minimum training is 40%. Fig. 3 illustrates the AUROC curves for the fraud detection problem with 40% minimum training with hyperparameter tuning. The detailed results of each ML method for fraud detection, such as computing time, false positive rate, false negative rate, and precision of 10-fold cross-validation, are available at https://osf.io/zpwuq. The reader can find comparative performance analysis of the methods with and without hyperparameter tuning. Based on the reported results, "PCA" and "Sampling" are consistently the methods with the highest Critical Ratio values with and without hyperparameter tuning. This result is mainly due to its much faster computational time than the other methods. Table 9 shows that hyperparameter tuning can improve precision on three supervised machine learning methods.

Table 9. Hyperparameter Impact in Supply Chain Fraud Detection Accuracy (10-fold)

| **ML method** | **With hyperparameter tuning** | **Without hyperparameter tuning** |
|---|---|---|
| Logistic Regression (LR) | 0.9186 | 0.8959 |
| XGBoost | 1 | 1 |
| LightGBM | 1 | 1 |
| Linear Discriminant Analysis | 0.9687 | 0.9740 |
| Naive Bayes ComplementNB | 0.9779 | 0.8639 |
| Random Forest (RF) | 1 | 1 |
| Random Forest-balanced | 1 | 1 |
| Balanced Bagging | 1 | 0.9999 |
| Balanced Random Forecast | 1 | 1 |
| Ensemble with Random Tree and LR | 0.9989 | 0.9990 |
| Ensemble with Decision Tree and AdaBoost | 1 | 1 |
| Ensemble with RF and RUSBoost | 0.9959 | 0.9959 |
| Neural Network-MLP | 0.9930 | 0.9854 |

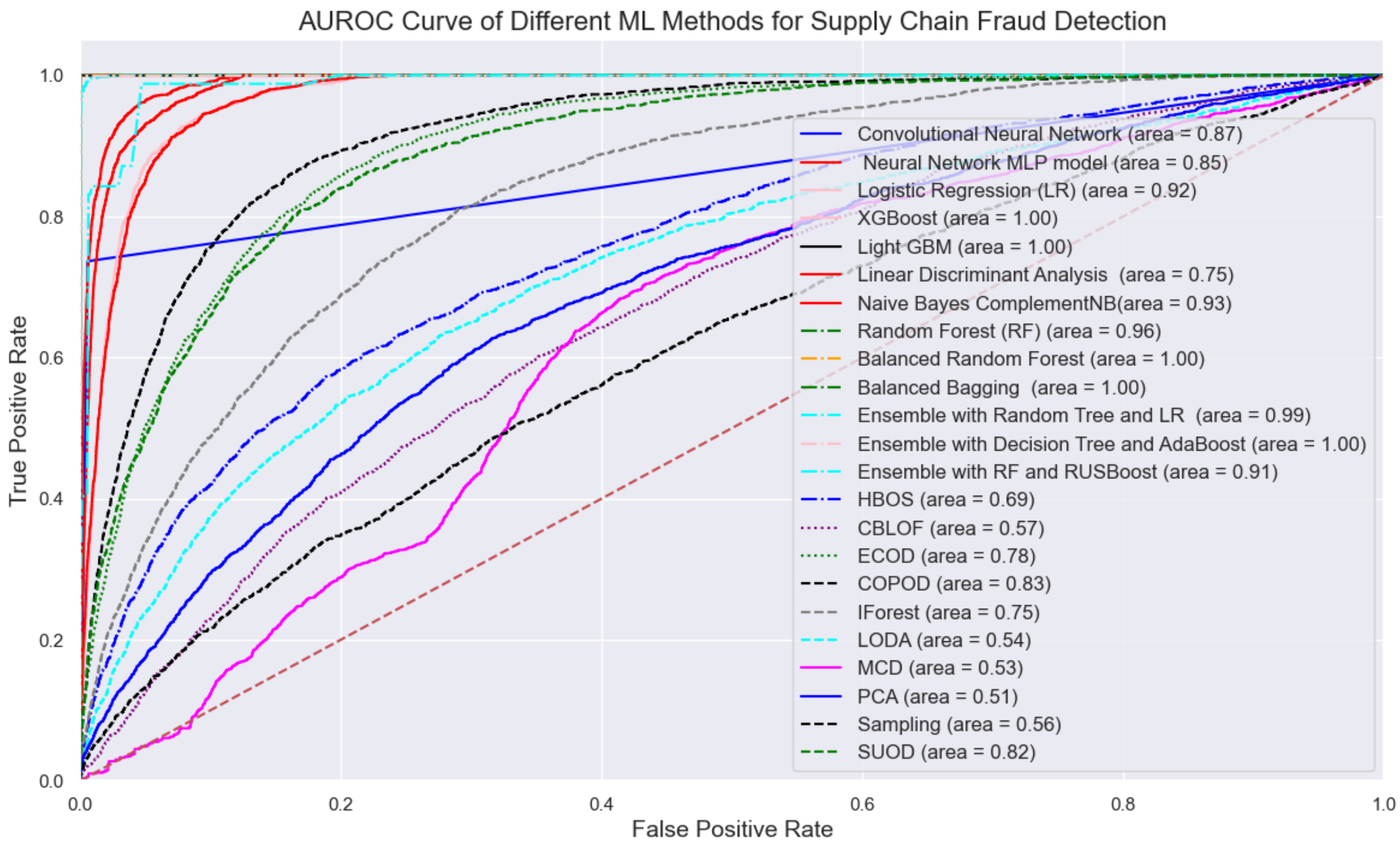

FIG. 3. AUROC Curve for supply chain fraud detection with hyperparameter tuning

*4.2. Machine Failure Detection*

This section provides the computational results for the machine failure detection problem. Failure to optimize the schedule of preventive maintenance leads to supply chain disruptions. A dataset that reflects real predictive maintenance encountered in the industry is used to test the ML algorithms. Preventive maintenance was addressed using real-world datasets with over 10,000 data points, where predictive algorithms like Neural Networks and ensemble methods demonstrated up to 99.9% accuracy. Hyperparameter tuning further improved the performance of these models, particularly in reducing false negatives, which is critical for avoiding costly downtimes. Integrating IoT data through interconnected sensors allows for real-time machinery monitoring, enabling preemptive actions and ensuring uninterrupted operations in technologically intensive supply chains. Fig. 4 presents the computational results with hyperparameter tuning. The detailed results of each ML method for machine failure detection, such as computing time, false positive rate, false negative rate, and precision of 10-fold cross-validation, are available at https://osf.io/zpwuq. The results indicate that Neural Network-MLP, Ensemble with Decision Tree and Adaboost, Random Forest, Naïve Bayes Complement NB, and Linear Discriminant Analysis methods outperform the other techniques in computational times with 99.5% accuracy when hyperparameter tuning is applied. Meanwhile, the 6 out of 24 method is 100% accurate when hyperparameter tuning is applied, as opposed to only 1 out of 24 when it is not used. Table 10 shows that hyperparameter tuning can improve precision on two supervised ML methods, MLP and CNN.

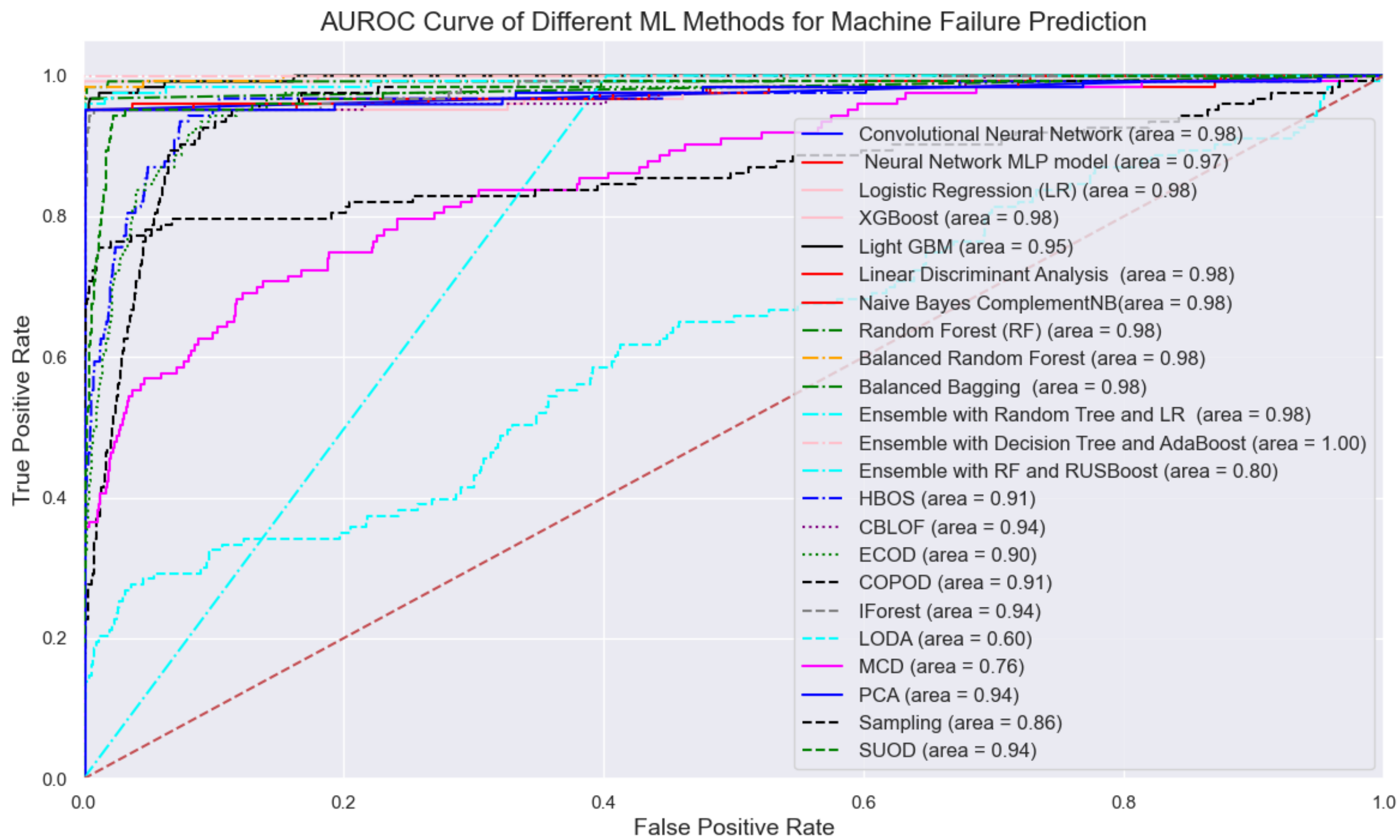

FIG. 4. AUROC Curve for machine failure detection with hyperparameter tuning

Table 10. Hyperparameter Impact in Machine Failure Detection Accuracy (10-fold)

| **ML method** | **With hyperparameter tuning** | **Without hyperparameter tuning** |
|---|---|---|
| Logistic Regression (LR) | 0.9913 | 0.9995 |
| XGBoost | 1 | 0.9995 |
| LightGBM | 0.9989 | 1 |
| Linear Discriminant Analysis | 0.9870 | 0.9995 |
| Naive Bayes ComplementNB | 0.9871 | 0.9995 |
| Random Forest (RF) | 0.9918 | 0.9995 |
| Random Forest-balanced | 1 | 0.9995 |
| Balanced Bagging | 1 | 0.9993 |
| Balanced Random Forecast | 1 | 0.9993 |
| Ensemble with Random Tree and LR | 1 | 0.9993 |
| Ensemble with Decision Tree and AdaBoost | 1 | 0.9995 |
| Ensemble with RF and RUSBoost | 0.7065 | 0.7065 |
| Neural Network-MLP | 0.9871 | 0.9995 |

*4.3. Material Backorder Prediction*

This section provides the computational results for the material backorder prediction problem. Material shortages, a common bottleneck in supply chains, were effectively addressed using ML models trained on inventory and sales data. Methods such as LightGBM and XGBoost achieved near-perfect accuracy (up to 100% with tuning) in predicting backorders. The models empower organizations to optimize inventory planning and improve customer service levels by identifying variables like inventory levels and lead times as critical features. This capability directly supports operational efficiency and profitability. The test dataset includes the variables that were listed in Table 5.

Fig. 5 presents the computational results for material backorder prediction with hyperparameter tuning. Sampling once again has the fastest computational times in this problem, while its accuracy is 93.4% when hyperparameter tuning is applied. The detailed results of each ML method for material backorder prediction, such as computing time, false positive rate, false negative rate, and precision of 10-fold cross-validation, are available at https://osf.io/zpwuq. Table 11 shows that hyperparameter tuning can improve the accuracy on nine supervised ML methods.

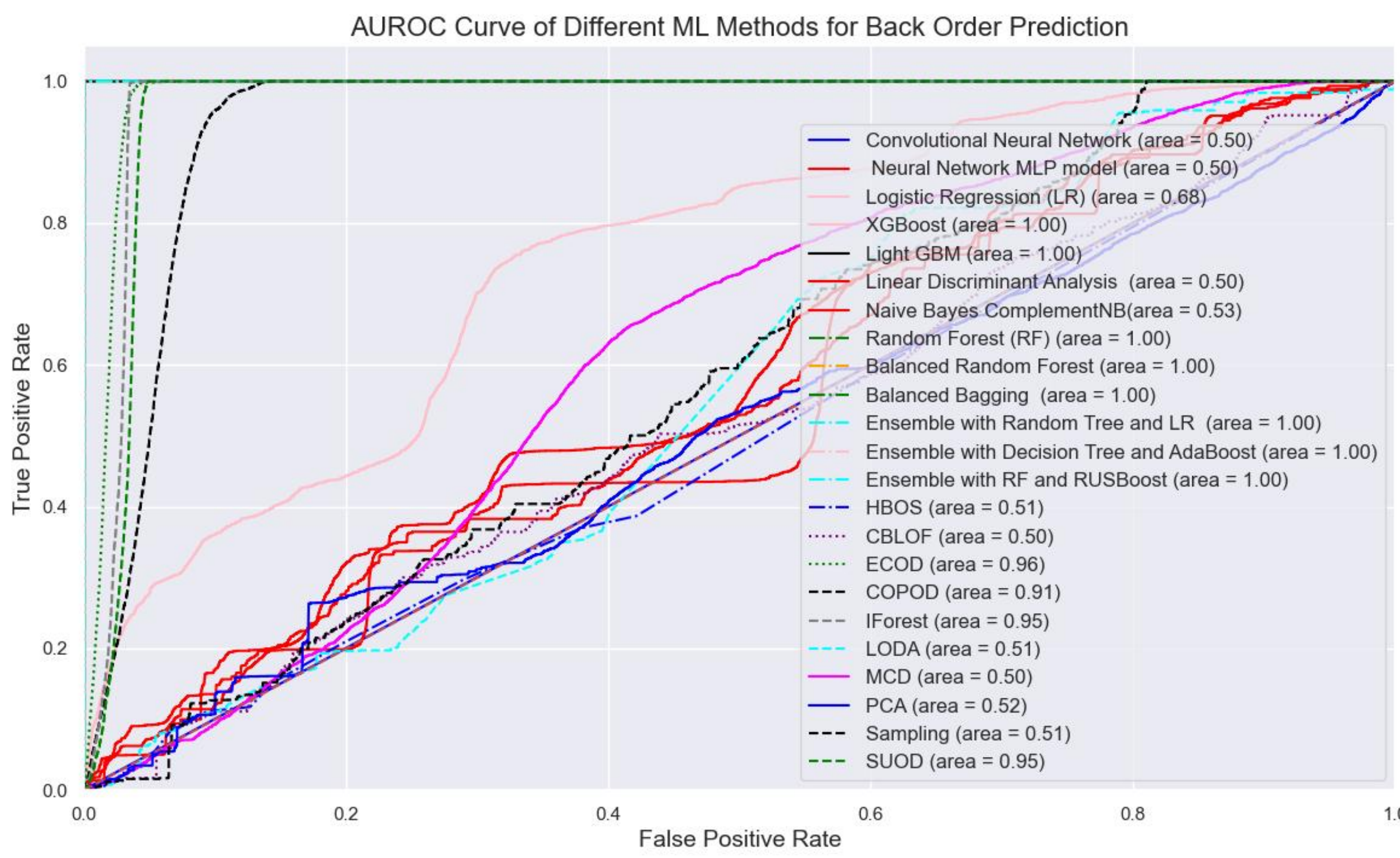

FIG. 5. AUROC Curve for material backorder prediction with hyperparameter tuning

Table 11. Hyperparameter Impact in Backorder Prediction Accuracy (10-fold)

| ML method | With hyperparameter tuning | Without hyperparameter tuning |
|---|---|---|
| Logistic Regression (LR) | 0.6487 | 0.6181 |
| XGBoost | 1 | 0.9912 |
| LightGBM | 1 | 0.9901 |
| Linear Discriminant Analysis | 0.9915 | 0.9905 |
| Naive Bayes ComplementNB | 0.8893 | 0.8569 |
| Random Forest (RF) | 1 | 0.9920 |
| Random Forest-balanced | 1 | 0.9904 |
| Balanced Bagging | 1 | 0.9916 |
| Balanced Random Forecast | 1 | 0.9064 |
| Ensemble with Random Tree and LR | 1 | 0.6431 |
| Ensemble with Decision Tree and AdaBoost | 1 | 0.9910 |
| Ensemble with RF and RUSBoost | 1 | 0.6771 |
| Neural Network-MLP | 0.9912 | 0.9912 |

## 5. Discussion

Every decision in the supply chain depends on some form of data, and while access to data is critical for supply chain visibility, using intelligent software to augment human decision makers has become necessary to take advantage of the vast amount of available data. Machine learning methods offer significant potential as a decision support system in various supply chain fields.

The AutoML framework developed in this study offers a systematic approach to data preprocessing, feature selection, model construction, and deployment. It stands out for its adaptability, scalability, and ease of integration into supply chain systems. By addressing fraud, maintenance, and backorder challenges with high precision, this framework reduces operational risks, minimizes financial losses, and improves overall supply chain resilience. These findings underscore the utility of machine learning as a critical enabler of data-driven decision-making in modern supply chains. Organizations adopting such frameworks are better positioned to navigate complexities, optimize resources, and achieve sustainable growth in increasingly dynamic global markets.

### *5.1. Classification measures*

This section provides a summary of results and a comparison between different metrics. Tables 12, 13, and 14 report the performance results of the algorithms with hyperparameter tuning for fraud detection, machine failure detection, and material backorder detection, respectively. The results without the hyperparameter tuning are also provided as a supplementary document.

While some supervised learning models achieved 100% precision in tasks such as material backorder prediction (as shown in Table 14), this level of performance raises concerns regarding potential overfitting or data leakage. Although extensive preprocessing steps were undertaken, including encoding, imbalanced class handling via SMOTE, and separation of training/testing sets, such perfect scores are atypical in real-world scenarios. Several factors could contribute to this phenomenon: (1) the predictive features may be overly correlated with the target variable, particularly in a synthetic or highly structured dataset; (2) SMOTE may have synthetically inflated the separability of classes; and (3) subtle data leakage (e.g., temporal leakage or label contamination during encoding) could have introduced artifacts that boosted model performance. Future work will incorporate stricter cross-validation protocols, leakage detection tests, and independent validation datasets to confirm model generalizability under realistic conditions and mitigate these risks.

Table 12. Performance comparison for supply chain fraud detection with hyperparameter tuning

| Method | Time (sec) | Correct Predicted | Overall Accuracy-10folds | ROC |
|---|---|---|---|---|
| Logistic Regression (LR) | 182.731 | 66771 | 0.9186 | 0.9222 |
| XGBoost | 14.224 | 72207 | 1 | 1 |
| LightGBM | 27.466 | 72207 | 1 | 1 |
| Linear Discriminant Analysis | 31.030 | 70414 | 0.9687 | 0.7488 |
| Naive Bayes Complement NB | 0.229 | 62172 | 0.9779 | 0.9289 |
| Random Forest | 177.836 | 72076 | 1 | 0.9607 |
| Random Forest-Balanced | 201.338 | 72206 | 1 | 1 |
| Balanced Bagging | 584.221 | 72207 | 1 | 1 |
| Balanced Random Forecast | 36.791 | 72200 | 1 | 1 |
| Ensemble with Random Tree and LR | 0.543 | 72022 | 0.9989 | 0.9876 |
| Ensemble with Decision Tree and AdaBoost | 195.562 | 72207 | 1 | 1 |
| Ensemble with RF and RUSBoost | 11.217 | 70016 | 0.9959 | 0.9077 |
| Neural Network-MLP | 193.688 | 71317 | 0.9930 | 0.8466 |
| Convolutional Neural Network | 229.556 | 71361 | 0.9980 | 0.8654 |
| COPOD | 4.790 | 65014 | 0.9088 | 0.8271 |
| HBOS | 0.299 | 53137 | 0.8951 | 0.6925 |
| CBLOF | 0.282 | 63525 | 0.8896 | 0.5669 |
| ECOD | 3.201 | 65256 | 0.9058 | 0.7781 |
| IForest | 2.414 | 54849 | 0.8994 | 0.7542 |
| LODA | 1.109 | 69919 | 0.9697 | 0.5404 |
| MCD | 123.324 | 60697 | 0.8846 | 0.5299 |
| PCA | 0.510 | 2790 | 0.8923 | 0.5059 |
| Sampling | 0.446 | 63545 | 0.8973 | 0.5591 |
| SUOD | 65.882 | 56465 | 0.9035 | 0.8210 |

Table 13. Performance comparison for machine failure detection with hyperparameter tuning

| Method | Time (sec) | Correct Predicted | Overall Accuracy-10folds | ROC |
|---|---|---|---|---|
| Logistic Regression (LR) | 6.483 | 3994 | 0.9913 | 0.9756 |
| XGBoost | 1.189 | 3994 | 1 | 0.9756 |
| LightGBM | 1.757 | 3988 | 0.9989 | 0.9512 |
| Linear Discriminant Analysis | 3.365 | 3994 | 0.9870 | 0.9756 |
| Naive Bayes Complement NB | 0.319 | 3994 | 0.9871 | 0.9756 |
| Random Forest | 4.757 | 3994 | 0.9918 | 0.9756 |
| Random Forest-Balanced | 5.450 | 3994 | 1 | 0.9756 |
| Balanced Bagging | 15.641 | 3996 | 1 | 0.9837 |
| Balanced Random Forecast | 7.483 | 3993 | 1 | 0.9794 |
| Ensemble with Random Tree and LR | 0.031 | 3994 | 1 | 0.9756 |
| Ensemble with Decision Tree and AdaBoost | 6.724 | 4000 | 1 | 1 |
| Ensemble with RF and RUSBoost | 0.418 | 2440 | 0.7065 | 0.7988 |
| Neural Network-MLP | 11.913 | 3993 | 0.9871 | 0.9715 |
| Convolutional Neural Network | 10.727 | 3994 | 0.9985 | 0.9756 |
| COPOD | 0.553 | 3690 | 0.9320 | 0.9128 |
| HBOS | 0.020 | 3733 | 0.9318 | 0.9065 |
| CBLOF | 0.053 | 3730 | 0.9337 | 0.9416 |
| ECOD | 0.068 | 3731 | 0.9322 | 0.9023 |
| IForest | 0.245 | 3733 | 0.9340 | 0.9419 |
| LODA | 0.150 | 3563 | 0.9148 | 0.6012 |
| MCD | 0.585 | 3640 | 0.9088 | 0.7568 |
| PCA | 0.017 | 3746 | 0.9350 | 0.9436 |
| Sampling | 0.020 | 3696 | 0.9303 | 0.8624 |
| SUOD | 17.201 | 3721 | 0.9342 | 0.9404 |

Table 14. Performance comparison for backorder detection with hyperparameter tuning

| **Method** | **Time (sec)** | **Correct Predicted** | **Overall Accuracy-10folds** | **ROC** |
|---|---|---|---|---|
| Logistic Regression (LR) | 232.668 | 234479 | 0.6487 | 0.6846 |
| XGBoost | 56.352 | 419430 | 1 | 1 |
| LightGBM | 81.662 | 419430 | 1 | 1 |
| Linear Discriminant Analysis | 7.902 | 415434 | 0.9915 | 0.5018 |
| Naive Bayes Complement NB | 1.727 | 361471 | 0.8893 | 0.5300 |
| Random Forest | 324.516 | 419430 | 1 | 1 |
| Random Forest-Balanced | 342.140 | 419430 | 1 | 1 |
| Balanced Bagging | 2123.180 | 419430 | 1 | 1 |
| Balanced Random Forecast | 198.406 | 419430 | 1 | 1 |
| Ensemble with Random Tree and LR | 4.615 | 419430 | 1 | 1 |
| Ensemble with Decision Tree and AdaBoost | 2482.109 | 419430 | 1 | 1 |
| Ensemble with RF and RUSBoost | 13.619 | 419427 | 1 | 1 |
| Neural Network-MLP | 182.448 | 415736 | 0.9912 | 0.5000 |
| Convolutional Neural Network | 627.362 | 415737 | 0.9916 | 0.5004 |
| COPOD | 19.973 | 383433 | 0.9066 | 0.9094 |
| HBOS | 4.293 | 376004 | 0.8964 | 0.5076 |
| CBLOF | 2.451 | 374574 | 0.8929 | 0.5043 |
| ECOD | 18.043 | 382970 | 0.9088 | 0.9562 |
| IForest | 10.745 | 381600 | 0.9088 | 0.9545 |
| LODA | 6.739 | 385992 | 0.9056 | 0.5080 |
| MCD | 115.532 | 369523 | 0.8929 | 0.4977 |
| PCA | 1.568 | 374439 | 0.8934 | 0.5175 |
| Sampling | 0.509 | 374672 | 0.8925 | 0.5134 |
| SUOD | 117.706 | 381686 | 0.9088 | 0.9546 |

### *5.2. Feature importance*

To understand which features contribute the most to predicting the target variable, the Shapley Additive Explanations (SHAP) method is used. This method is utilized to add further transparency to the ML methods. SHAP values indicate how much impact each factor had on the target variable. SHAP methodology enables the prioritization of features that determine the classification and prediction through machine learning models (Perez and Bajorath, 2020). It should be noted that SHAP values depend on the specific set of observations. They will change as the observed data changes.

The SHAP summary bar plot in Fig. 6 visualizes the global feature importance for the Random Forest model trained on the supply_chain_fraud_detection dataset. The x-axis represents the mean(|SHAP value|), the average absolute impact a feature has on the model's output magnitude across all predictions. Features are ranked on the y-axis in descending order of importance. This plot reveals that Type is the most influential predictor, possessing the highest mean absolute SHAP value. The stacked bars, colored by class, indicate that Type's importance is predominantly driven by its contribution to predicting Class 1 (red bar), suggesting specific Type values are potent indicators of the positive class (e.g., fraud). In contrast, features like feature_22 and Late_delivery_risk demonstrate significant, more balanced contributions to Class 0 and Class 1 predictions.

The SHAP force and decision plots (Figs. 7 and 8) provide a local explanation for a single prediction from the model. They illustrate how individual feature values for one instance work together to shift the model's output from the base value (the average prediction over the dataset) to the final output score, f(x). For the specific instance shown, the model outputs a high score of f(x) = 1.00, strongly predicting Class 1. The red segments represent features with positive SHAP values that pushed the prediction higher (towards Class 1). The plots clearly show that the Type feature provided the largest positive contribution, acting as the primary driver for this prediction. Other features like Shipping Mode and Department Name contributed positively, cumulatively overcoming any minor negative contributions from different features to result in a confident Class 1 prediction.

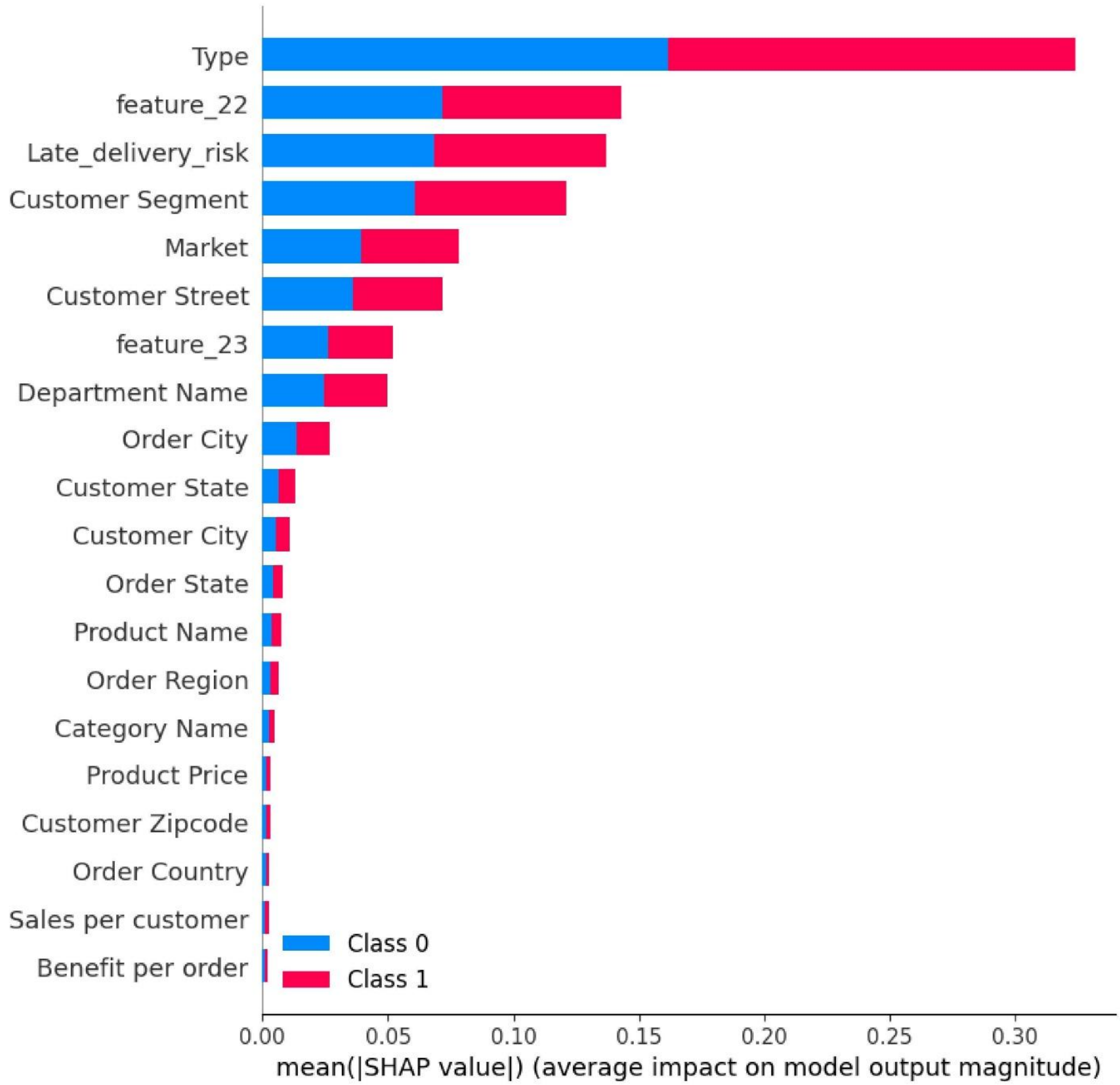

FIG. 6. SHAP summary bar plot

Fig. 7 illustrates the SHAP values of each variable. The results show that “Type” with a SHAP value of 0.081 contributed the most towards detecting the fraudulent activities. Another way of showing the contribution of each variable to the prediction of the target variable is the force plot, as provided in the supplemental file. Negative SHAP values are displayed on the left side of Fig. 8, while the positive values are on the right.

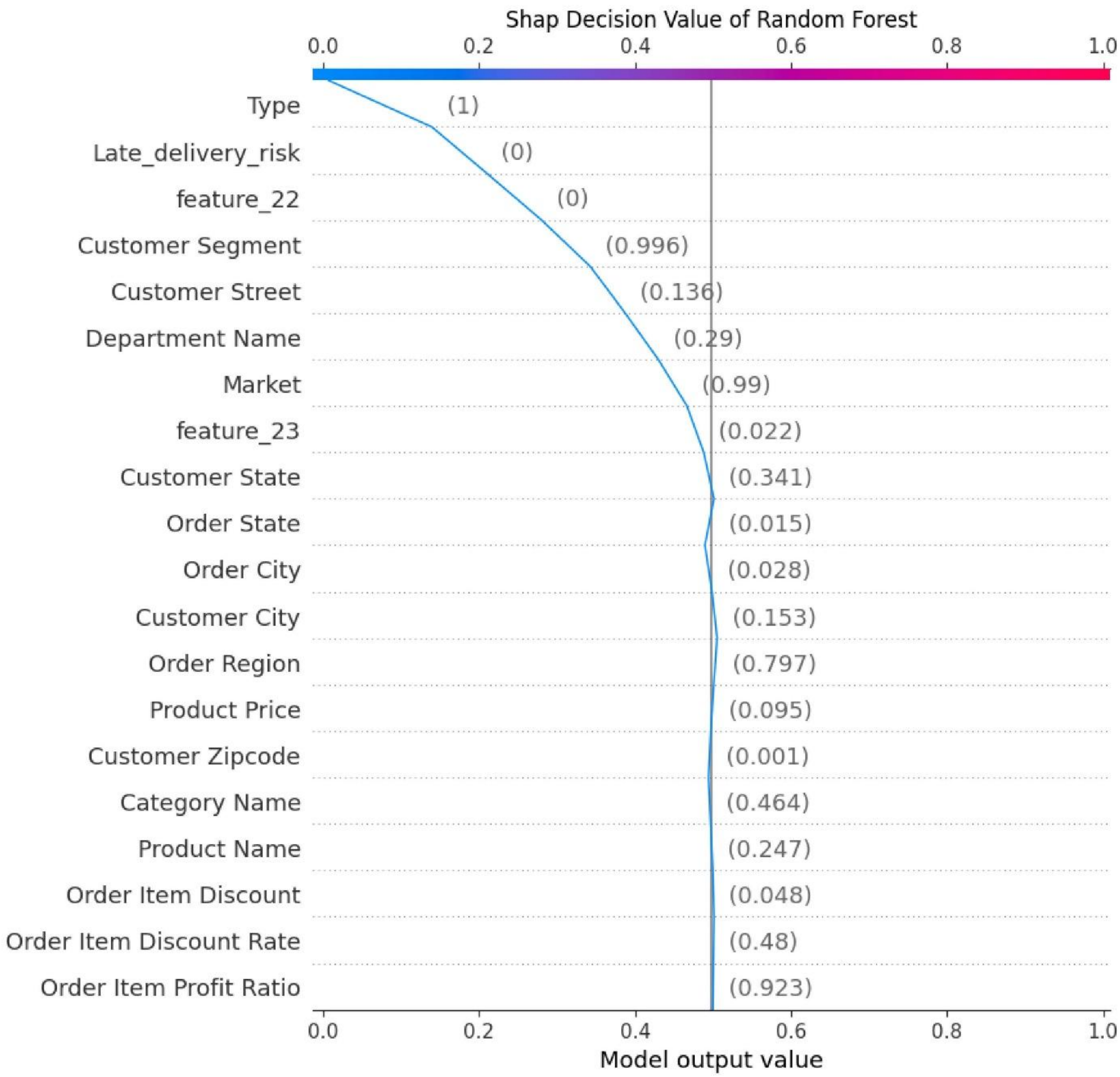

FIG. 7. SHAP values on the importance of features in the supply chain fraud dataset

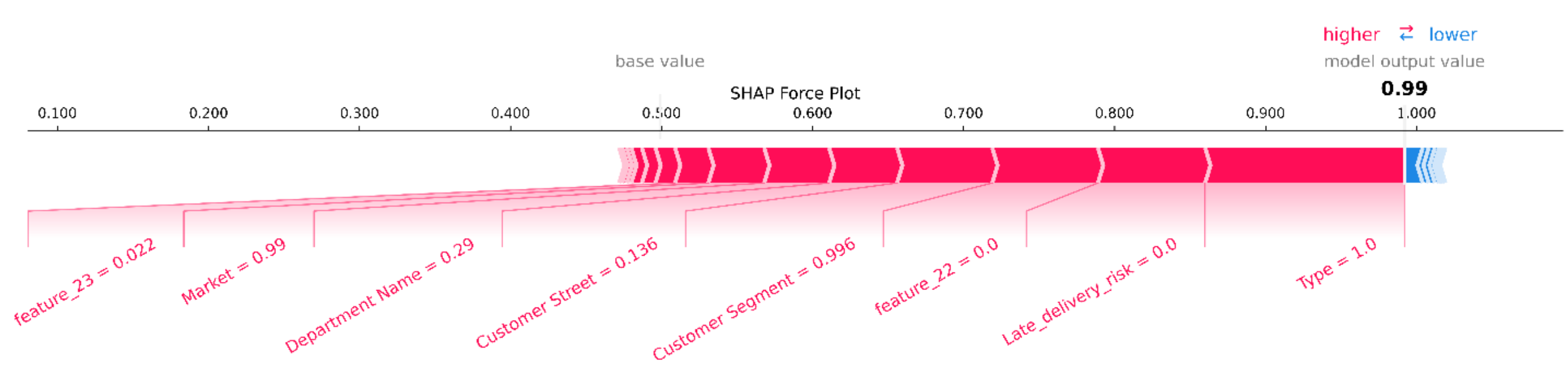

FIG. 8. Force Plot of SHAP values on the importance of features in the supply chain fraud dataset

This chart explains why the model arrived at a final high-risk score of 0.99 for this specific event, which is much higher than the average outcome of about 0.50. The model started with the average and then adjusted its score based on the details of this particular case. Several factors, shown in red, pushed the score up. The most significant reason for the high score was that the Type of event was 1.0. Other details, like the Late_delivery_risk and Customer Segment, also strongly contributed to the increase. While one small factor, shown in blue, tried to lower the score slightly, its effect was insignificant. Combining these red factors, especially the Type, made the model confident in its high-risk prediction.

Compared to earlier studies, which focus on a specific supply chain application or algorithm, our framework offers a holistic approach that automates model selection, encoding strategy, and hyperparameter optimization. Integrating semi-supervised and unsupervised methods—a rarely implemented practice in supply chain research—enables learning from partially labeled or unlabeled data, addressing the real-world challenge of limited annotation. Moreover, our extensive benchmarking across encoding methods and ML models reveals nuanced performance trade-offs not discussed in previous work. This makes the proposed framework both technically innovative and practically scalable for various supply chain security challenges.

As shown in Table 15, encoding methods significantly impact model performance. Leave-One-Out encoding consistently outperformed others across all three datasets, particularly in fraud detection and preventive maintenance. In contrast, One-Hot and Hashing—while computationally efficient—yielded lower predictive accuracy, likely due to their inability to capture complex relationships between categorical features and the target variable. These findings suggest that supervised encodings with label-awareness (e.g., Leave-One-Out, James-Stein) are more effective when dealing with structured, high-cardinality supply chain data.

Table 15. Impact of Encoding Methods on Model Performance (Accuracy %)

| **Encoding Method** | **Target Label Needed** | **Fraud Detection** | **Material Backorder** | **Preventive Maintenance** |
|---|---|---|---|---|
| Hashing | No | 84.2% | 86.7% | 89.1% |
| Count | No | 85.5% | 87.3% | 90.2% |
| Leave-One-Out | Yes | **88.0%** | **89.3%** | **93.4%** |
| James-Stein | Yes | 86.1% | 88.5% | 91.7% |
| One-Hot | No | 83.7% | 84.5% | 85.0% |
| M-Estimate | Yes | 87.4% | 88.9% | 91.9% |

Note: Performance was measured using XGBoost with tuned hyperparameters and SMOTE to measure class imbalance.

## 6. Conclusions

One of the popular terms used in the supply chain industry is resilience. Compounded by the global pandemic and political tensions in various regions, restoring resiliency and predictability to the supply chain system is at the top of the agendas of industrial leaders. Reducing shortages through the prediction of backorders is another area with great emphasis. Association of Certified Fraud Examiners (ACFE) reports that only 39.2% of the fraud cases surveyed were discovered by management review, internal audit, document examination, or account reconciliations, while 44.7% were found by accident. Similarly, the Economist Intelligence Unit report indicates that only 30% of supply chain executives were actively addressing corruption and bribery in their supply chains. The results imply both the need for fraud detection mechanisms and the opportunity for significant savings that can be obtained from such mechanisms for the supply chains.

A significant advantage of data analytics while examining supply chain transactions is the quick detection of fraud. Fraudulent transactions that may have gone unnoticed for an extended period may now be uncovered soon after they occur. Sophisticated algorithms can now go through thousands of third-party vendor invoices and work orders to reveal fraud instances in supply chains, including overcharging, double billing, and theft. This translates to significant savings for supply chains, especially the ones with complex global networks where there are a large number of suppliers to keep track of. In many cases, fraud can be detected before an invoice is paid, rather than years of audits revealing the problem. Another area where supply chain disruptions can stem from is misplaced maintenance activities. Predictive maintenance is about using sensor data to foresee the need for maintenance before it occurs, thus reducing the potential downtimes. Data availability, coupled with recent developments in computational power, has paved the way for machine learning methods to emerge as an efficient tool in combating supply chain disruptions.

This study demonstrates the potential of automated machine learning (AutoML) frameworks to enhance supply chain security through advanced data analysis. Key findings indicate the exceptional performance of specific ML techniques in solving critical supply chain problems. Fraud detection, machine failure detection, and material backorder prediction achieved near-perfect accuracy with several methods, significantly reducing prediction errors. Hyperparameter tuning notably enhanced precision across supervised ML models, with techniques such as LightGBM, Random Forest, and Ensemble Methods achieving up to 100% precision in some instances.

Integrating semi-supervised and unsupervised methods provided a robust approach for handling unlabeled data, demonstrating their importance in real-time, large-scale supply chain applications. Moreover, using SHAP values highlighted feature importance, enhancing transparency and trustworthiness of predictions, a crucial aspect for decision-makers in complex supply chain environments.

The proposed AutoML framework streamlines data preprocessing, feature selection, and model optimization, making it scalable and adaptable to diverse supply chain challenges. Its application has proven invaluable in addressing fraud detection, predictive maintenance, and material backorder forecasting, with practical implications for enhancing operational efficiency, reducing disruptions, and minimizing costs. This research underscores the utility of machine learning as a transformative tool for modern supply chains, offering actionable insights and a pathway to resilient, data-driven decision-making.

Strengthening supply chain resilience is a top priority, and our research demonstrates that automated machine learning frameworks are a powerful tool for achieving this. This study successfully applied these techniques to achieve high accuracy in detecting fraud, predicting machine failures, and forecasting material backorders, using methods that make the models' reasoning transparent and trustworthy for decision-makers. However, deploying such a framework in a live business environment requires addressing key practical challenges. Specifically, the model must be fast enough to make predictions in real-time, it must integrate smoothly with existing company software like financial and inventory systems, and it needs a process for continuous retraining to remain effective as supply chains and security threats evolve. Acknowledging these real-world hurdles, this work provides a technical proof-of-concept and a practical pathway toward building more secure, efficient, and data-driven supply chains.

**Funding Statement**

This research received no specific grant from any funding agency in the public, commercial, or not-for-profit sectors.

## Appendix

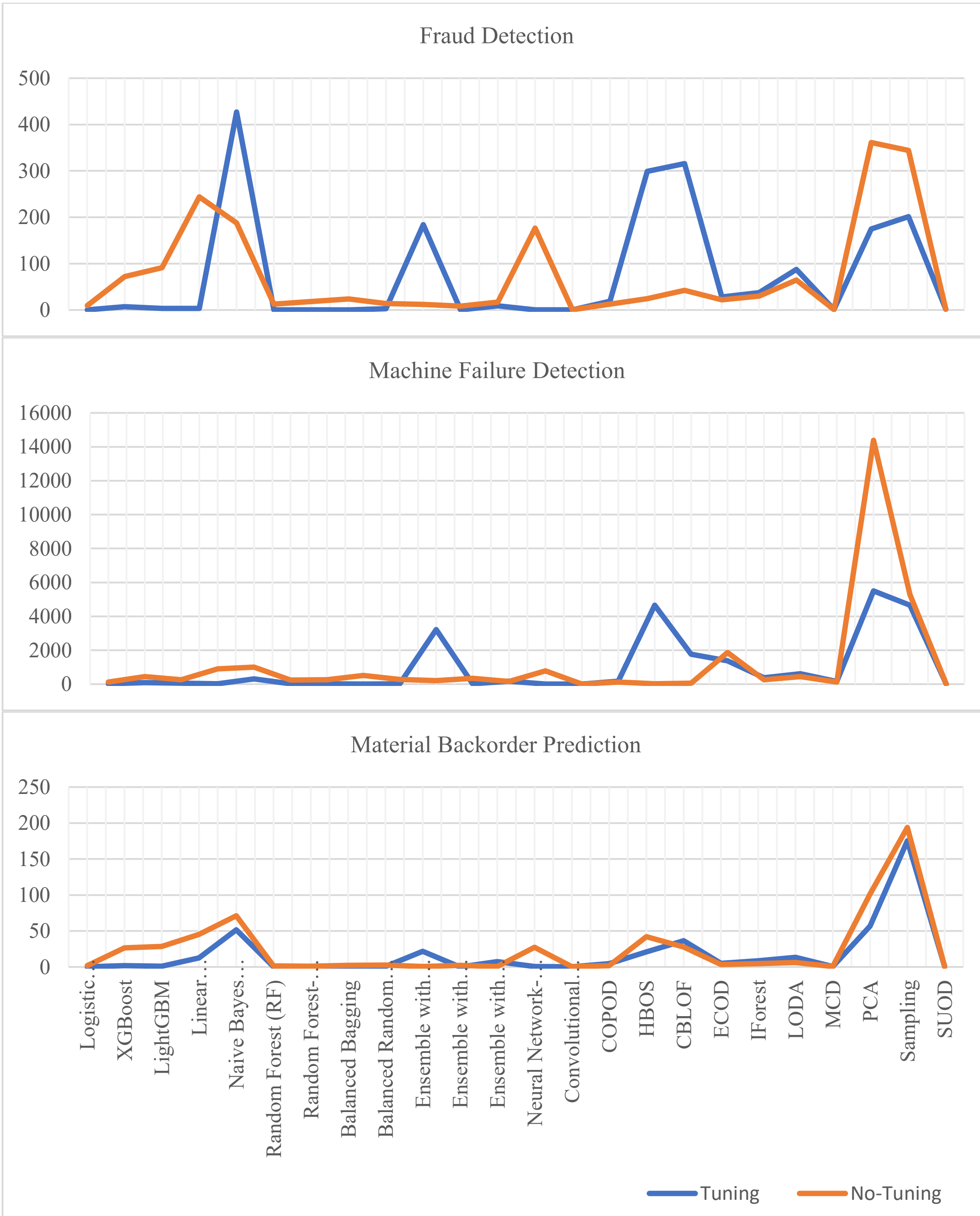

FIG A1. Critical Ratios of ML methods

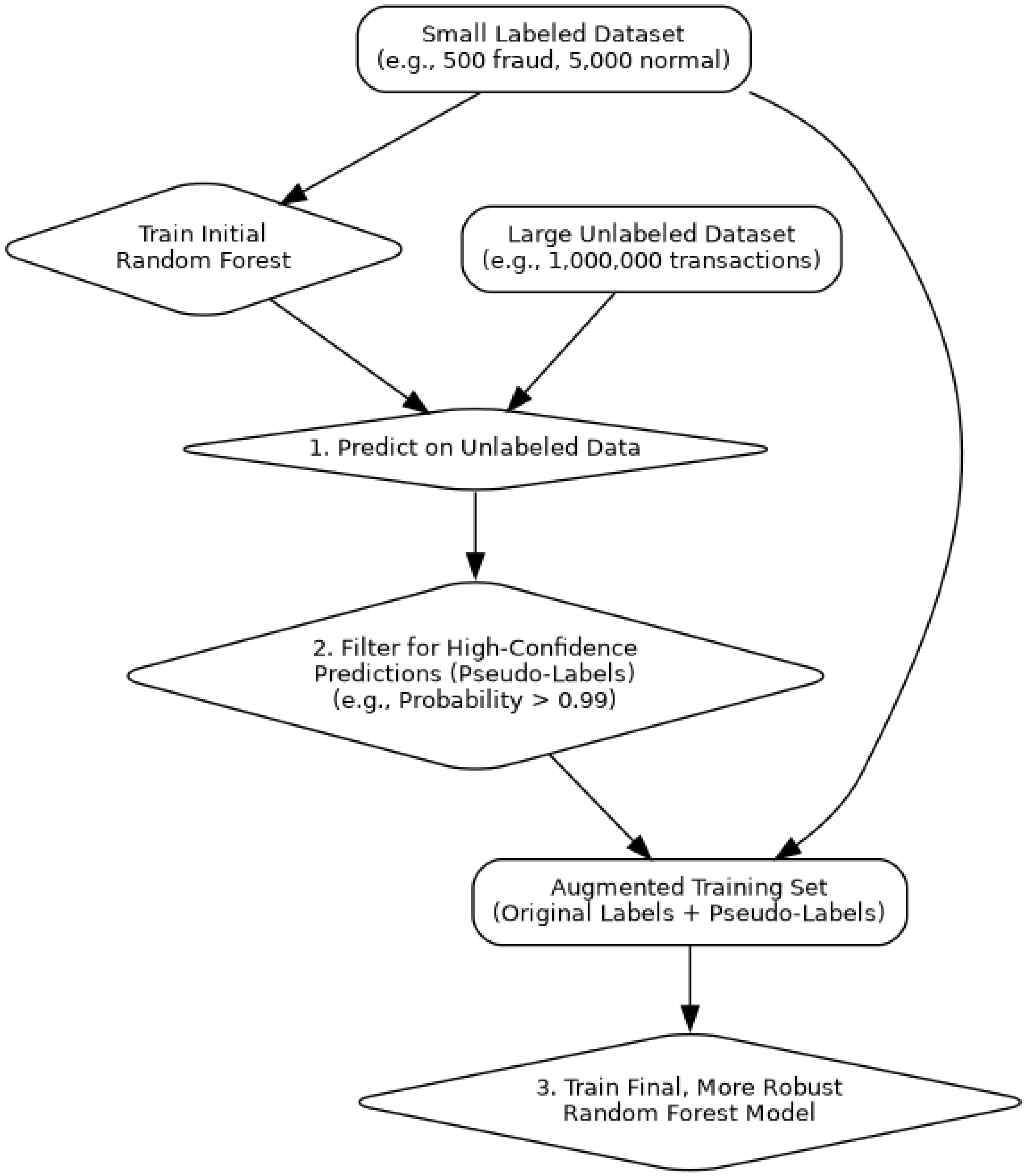

FIG A2. Illustration of Semi-supervised learning

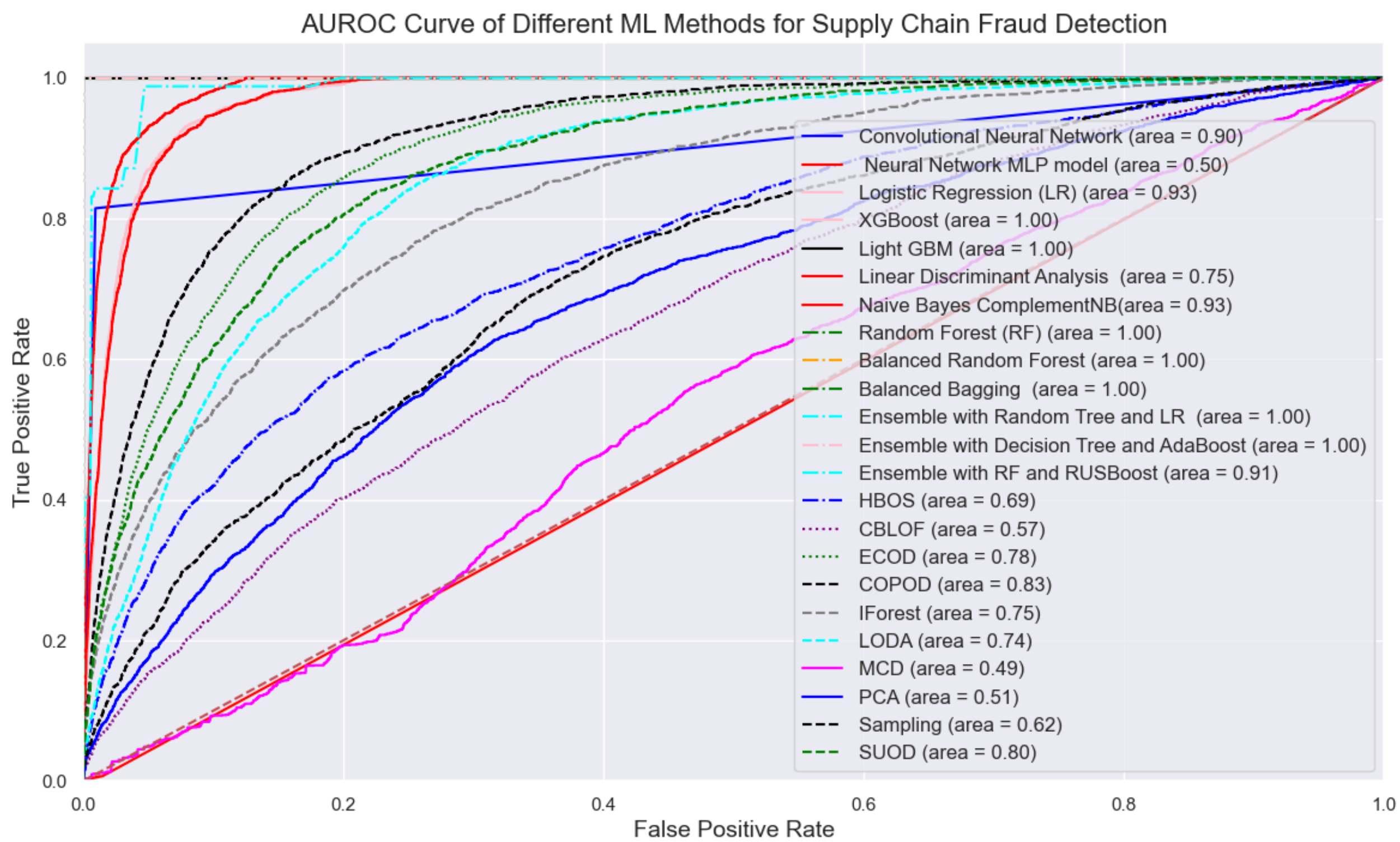

FIG. A3. AUROC Curve for supply chain fraud detection without hyperparameter tuning

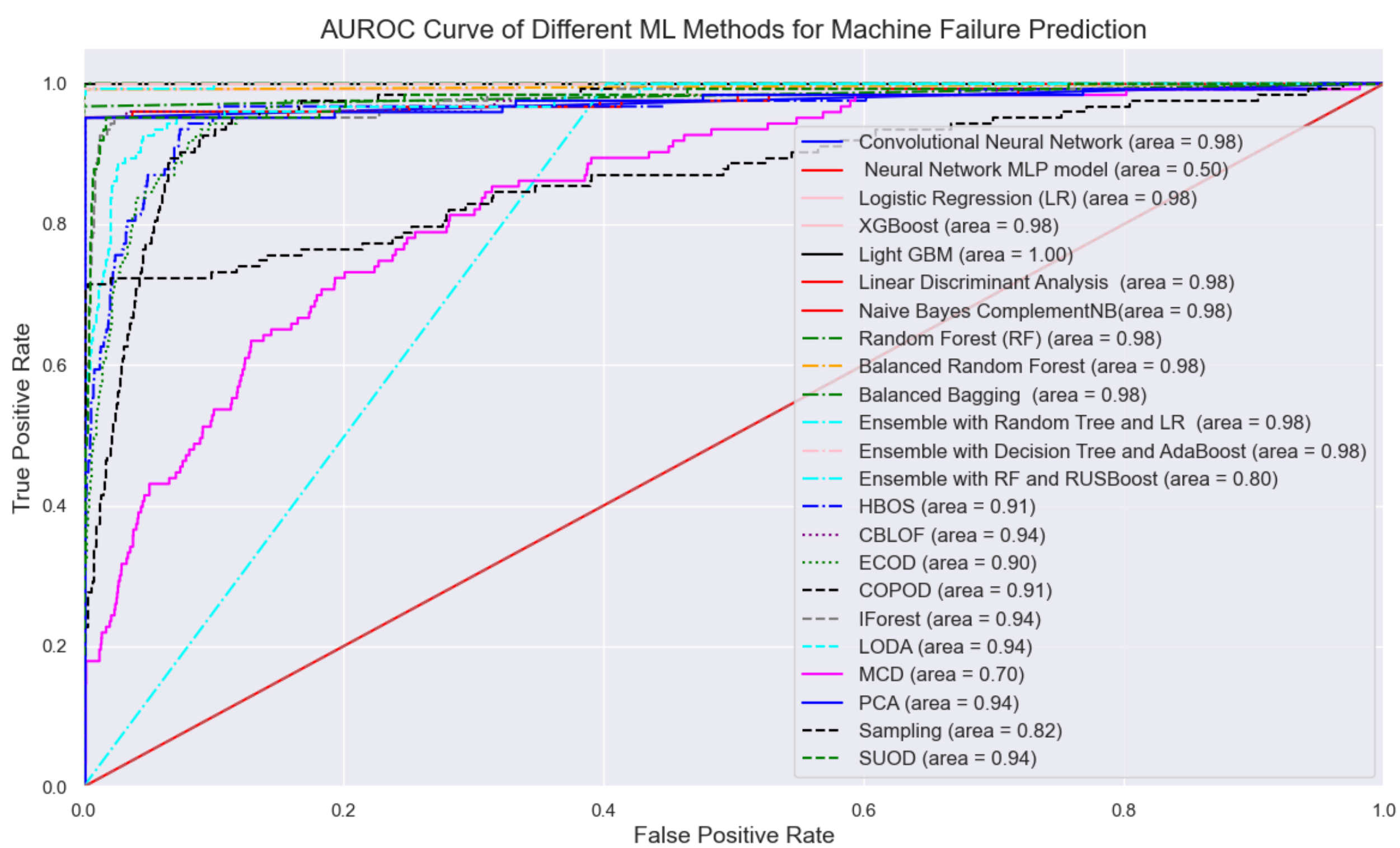

FIG. A4. AUROC Curve for machine failure detection without hyperparameter tuning

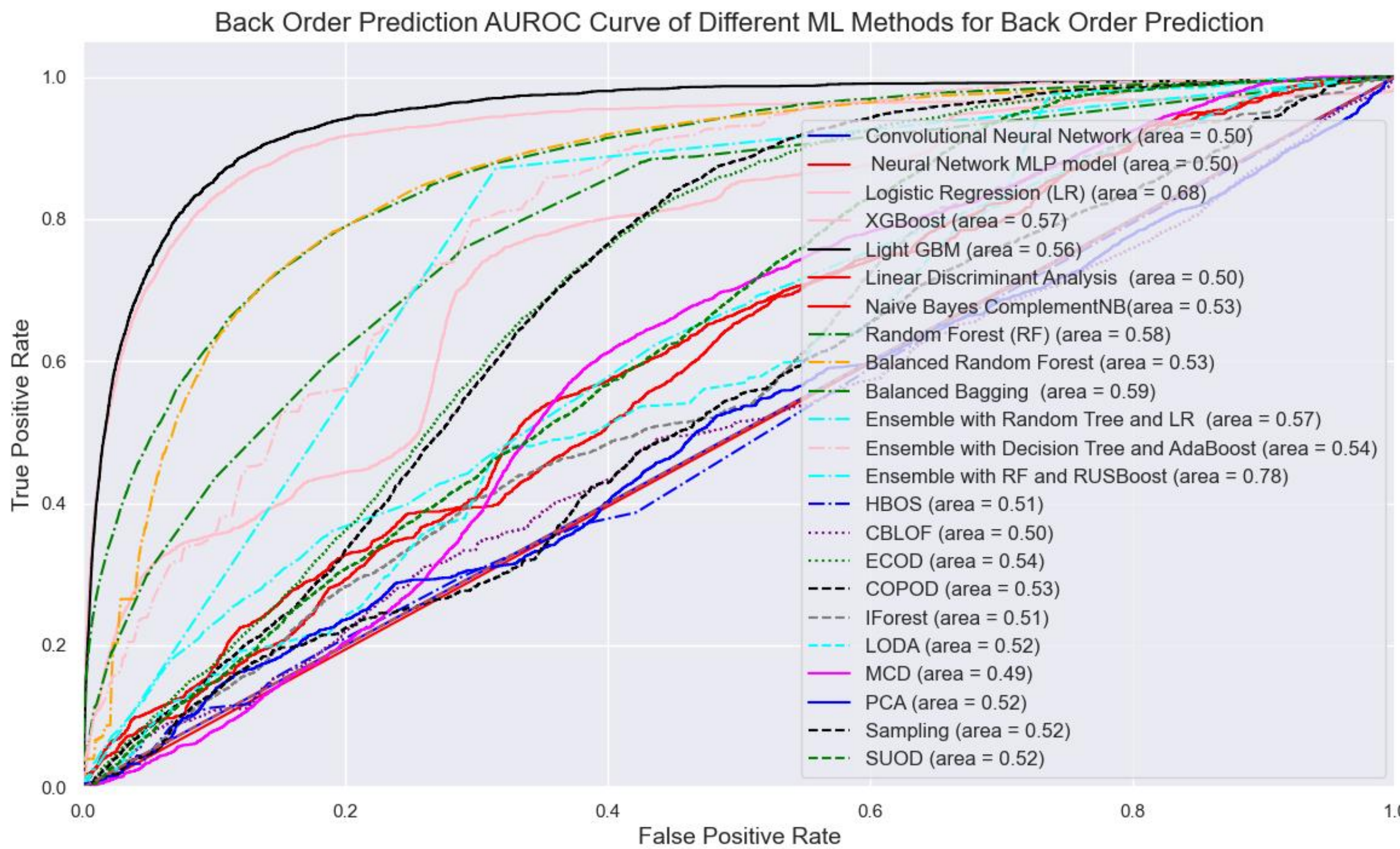

FIG. A5. AUROC Curve for material backorder prediction without hyperparameter tuning